\documentclass[final,1p,times]{elsarticle}
\usepackage{etoolbox}
\patchcmd{\subsubsection}{\itshape}{\bfseries}{}{}
\patchcmd{\subsection}{\itshape}{\bfseries}{}{}

\usepackage{amssymb}
\usepackage{amsmath}
\usepackage{amsthm}
\usepackage{booktabs}
\usepackage{bm}
\usepackage{graphicx}
\usepackage{tabularx}
\usepackage{makecell}
\newtheorem*{proposition}{Proposition}
\theoremstyle{remark}
\newtheorem*{remark}{Remark}

\journal{Computer Methods in Applied Mechanics and Engineering}

\begin{document}

\begin{frontmatter}

\title{A Constitutive Markov Physics-Informed Neural Operator (MPNO) for Autoregressive Stability in Transient Dynamics}

\author[1]{Wenpu Du}
\author[1]{Peng Zhou}
\author[1]{Yunlong Xia}
\author[1]{Sinuo Xin}
\author[1]{Congcong Zhang}
\author[1]{Boyang Zhang}
\author[1]{Yi Zhang}
\author[1]{Wenzheng Xu\corref{cor1}}
\ead{xuwznuc@163.com}

\cortext[cor1]{Corresponding author}

\affiliation[1]{organization={School of Environment and Safety Engineering, North University of China},
	city={Taiyuan},
	postcode={030051},
	country={China}}

\begin{abstract}

Applied to transient-dynamics PDEs with strong discontinuities, neural operators expose an autoregressive instability: in concrete-penetration stress-field prediction, the wavelet neural operator (WNO) diverges to infinity in autoregressive rollout, while MeshGraphNets collapse to zero predictions because the sparse stress signal is difficult to train on. By excluding numerous engineering factors one by one, WNO's instability is traced to the lack of a structural constraint on the spectral radius of its effective propagation operator. The Fourier neural operator (FNO) remains stable in the present measurements, yet it too lacks this constraint: its stability is emergent from random initialization and training trajectory rather than constructively guaranteed. A constitutive Markov physics-informed neural operator (MPNO) is therefore proposed, modeling one-step evolution as a Markov (row-stochastic) propagation operator. Physics-coupled edge weights (acoustic-impedance harmonic mean, contact area, and traction amplitude) encode material-interface constitutive information into a nonnegative symmetric adjacency matrix $W$; after normalization by $\lambda_{\max}$ of the graph Laplacian $L=D-W$ estimated online via Rayleigh-quotient power iteration, the propagation operator $P=I-\alpha\tilde{L}$ is constructively constrained to spectral radius $\rho(P)\le1$, suppressing exponential amplification of autoregressive errors by mechanism. Stability is thus engineered as a designable architectural property of the linear propagation operator (constructively nonexpansive by construction, with full-model autoregressive stability confirmed empirically across seeds) rather than an optimized loss objective, in contrast to the PINN/PINO route that imposes physics as soft PDE-residual constraints. Applicability is demonstrated on three PDEs: on time-evolution problems (Burgers, and two-dimensional transverse-section concrete penetration), MPNO rolls out stably with bounded error over the entire active horizon on all test seeds at impact speeds of 100/135/165 m/s, with stability a constructive spectral guarantee rather than an emergent outcome of training (WNO diverges everywhere; FNO is stable in practice but offers no such guarantee); the single-step relative $L_2$ error is $0.7304\pm0.0008$ (better than WNO, comparable to FNO's single-step error, at about one quarter of FNO's parameters). The edge-weight formula transfers across scenarios by replacing material-property variables. The model has about 20K parameters, delivering roughly $10^5\times$ inference speedup over LS-DYNA.
\end{abstract}

\begin{highlights}
\item The constitutive Markov physics-informed neural operator (MPNO) is proposed, encoding autoregressive stability as a constructive spectral guarantee $\rho(P)\le 1$ on the linear propagation operator rather than an optimized training objective, with full-model stability confirmed empirically across seeds.
\item Physics-coupled edge weights (acoustic-impedance harmonic mean $\times$ contact area $\times$ traction amplitude) embed material-interface constitutive information into graph propagation, transferring across PDE scenarios by replacing material-property variables alone.
\item Stable, bounded-error autoregressive rollout on the evaluation test seeds of three PDEs (concrete penetration, Burgers, and Darcy) with only about 20K parameters; the stability guarantee further transfers to an out-of-distribution two-dimensional compressible-Euler shock tube, where MPNO stays stable over all 91 rollout steps while FNO and WNO diverge (a single shock-tube trajectory).
\item Ablations and $\lambda_{\max}$ extrapolation stress tests show the stability contribution of spectral normalization is metric-independent: no\_spec's emergent stability breaks monotonically as the amplification factor grows.
\end{highlights}


\end{frontmatter}

\section{Introduction}
\label{sec:1}

\begin{figure}[!ht]
	\centering
	\includegraphics[width=\linewidth]{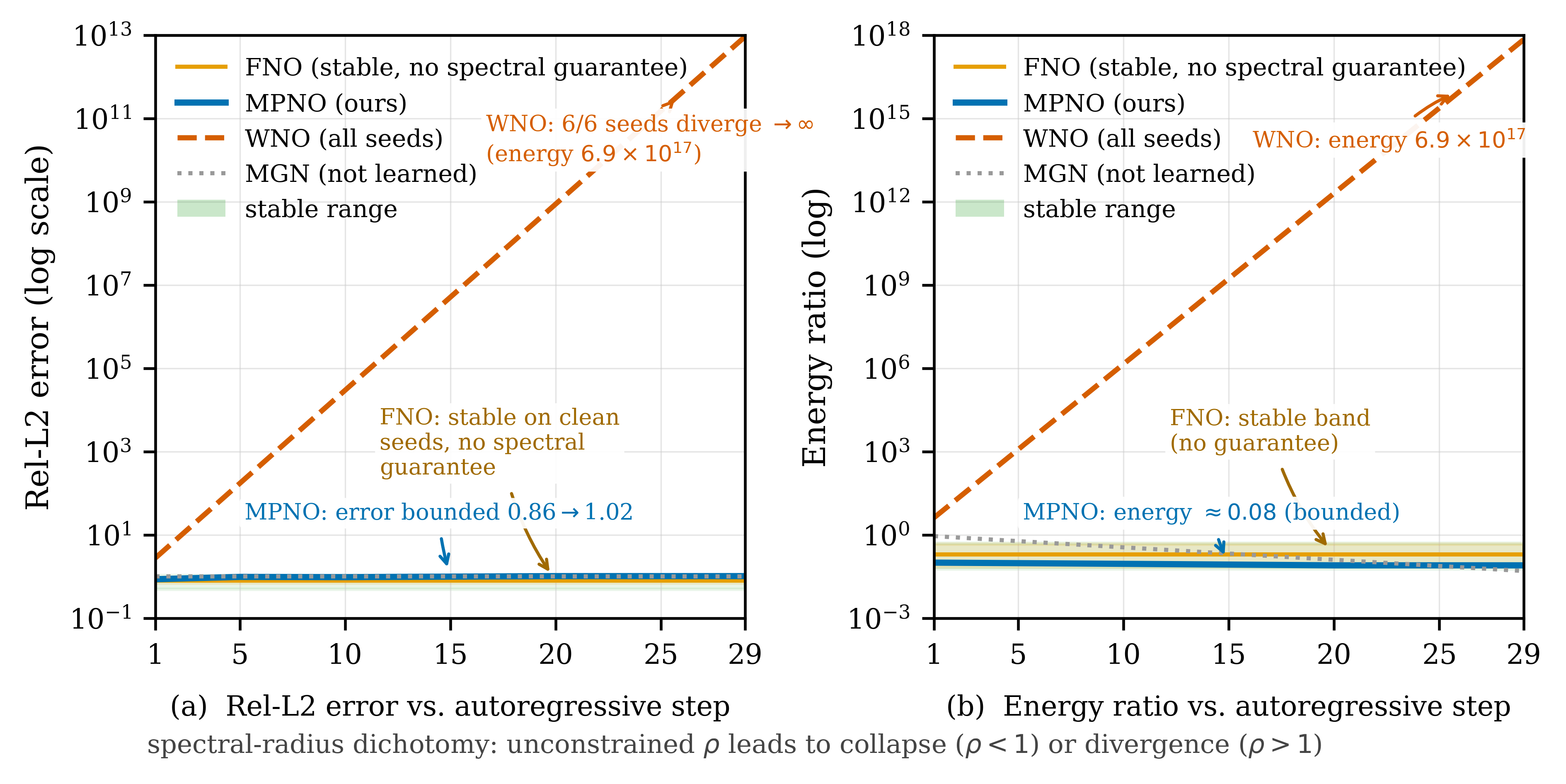}
	\caption{Autoregressive rollout of neural operators on concrete-penetration stress-field prediction (29 steps, frames 11--39 of the 40-frame window; schematic, anchored to measured single-step errors and end-of-rollout error bounds). MPNO (solid blue) stays bounded throughout; WNO (red dashed) diverges to infinity on all seeds; MeshGraphNets (gray dotted) fails to learn (error stays near 1); FNO (orange band) remains stable on the clean test seeds but, unlike MPNO, carries no spectral guarantee.}
	\label{fig:1}
\end{figure}

The starting point is a striking failure: the wavelet neural operator (WNO) \cite{ref1}, trained to convergence, collapses when driven in autoregressive rollout. WNO is a spectral operator proposed by Tripura and Chakraborty \cite{ref1} that replaces FNO's Fourier basis with wavelets for time--frequency localization; its 3D variant (WNO3D) was applied to concrete-penetration data and its 2D evaluation configuration (WNO2D) trained to convergence. Under single-step (teacher-forced) prediction, the 2D WNO attains a relative $L_2$ error of 0.7422, higher than the proposed model's $0.7304\pm0.0008$. When the teacher signal is removed and the model is fed its own predictions, the autoregressive rollout diverges to infinity on every test seed, the physical field degenerating into numerical noise within a few time steps. In contrast, FNO attains 0.7210 (slightly better than the proposed model's) and remains stable on all clean test seeds, yet its spectral radius is unconstrained and its stability is not constructively guaranteed. Fig.~\ref{fig:1} compares the autoregressive rollouts over the full data horizon.

This failure initially resembles overfitting, undertraining, or insufficient data. To rule these out, more than thirty engineering factors were varied one by one (learning rate from $10^{-4}$ to $10^{-2}$, batch size from 4 to 32, network depth from 2 to 8 layers, hidden width from 16 to 128, plus Sobolev losses, spectral normalization, and scheduled sampling), none of which prevented the collapse. The conclusion is clear: \textbf{this is not a hyperparameter problem but an architectural one.} MPNO, the answer proposed here, is precisely an architecture that writes stability into the spectral structure of the propagation operator, making bounded autoregressive error a constructive property.

The mechanism is captured by the dichotomy of the spectral radius (full argument in \S\ref{sec:3_3}): for any linear or locally linearized autoregressive system whose spectral radius is unconstrained, the long-term behavior is strictly dichotomous: errors grow exponentially when $\rho>1$ (catastrophic divergence) and decay exponentially when $\rho<1$ (excessive dissipation; the predicted field smooths toward uniformity over long horizons). Whether rollout stays stable therefore hinges on $\rho$ falling in a shrinking band near the critical value, determined entirely by the chance of initialization and training trajectory, with no structural guarantee. This is a conditional statement: it characterizes what unconstrained spectral radii entail, not a claim that FNO or WNO collapses under every configuration. The error blow-up observed for WNO on the same data is consistent with this dichotomy; the energy collapse of MGN stems from underfitting (single-step error near 1), a distinct phenomenon.

An alternative is to append physical constraints to the loss so that the model learns to stay stable. This is the approach of physics-informed neural networks (PINNs) \cite{ref2} and their operator variants (PINOs) \cite{ref3}, which cast momentum and energy conservation as PDE residuals penalized in the loss. On PDEs with strong discontinuities this route faces structural difficulties: the composite PINN loss (data term plus pointwise residual term) exhibits ill-conditioned gradient flows whose terms are badly scaled, requiring specialized mitigation to converge \cite{ref4}, and PINNs attain limited accuracy relative to classical methods on shocked/intermittent flows \cite{ref5}. Early internal comparisons reproduced the observation: residual and data-loss gradients were irreconcilably disparate in magnitude, delivering conflicting signals at every optimization step and reducing training to a zero-sum trade-off between data fitting and physics satisfaction. Pointwise residual constraints thus face a systematic gradient-conditioning obstacle on strongly discontinuous PDEs, although this judgment is limited to pointwise residual forms (see \S\ref{sec:3_3_4}).

The joint failure of both routes points to a common root cause: \textbf{stability cannot be an objective optimized in the loss; it must be a property constructed in the architecture of the propagation operator.} The constitutive Markov physics-informed neural operator (MPNO) proposed here rests on this principle. Its boundary with PINN/PINO is clear: MPNO's stability goal never enters the loss and generates no physical-residual gradient; it is carried by the spectral structure of the propagation operator $P$. The core mechanism consists of three coupled components:

(i) an adjacency matrix $W$, given by the product of the acoustic-impedance harmonic mean, contact area, and traction amplitude (together with a zero-initialized MLP correction); its nonnegative symmetry directly determines positive semidefiniteness of the graph Laplacian $L=D-W$;

(ii) a Markov propagation operator $P=I-\alpha\tilde{L}$, whose spectral radius is normalized against $\lambda_{\max}$ estimated online via Rayleigh-quotient power iteration and thus constrained to $\rho(P)\le1$, keeping autoregressive rollout errors bounded in practice;

(iii) the Rayleigh-quotient power iteration, which estimates $\lambda_{\max}$ at $O(E)$ cost so that spectral normalization is feasible within the training loop. The three components form a one-way dependency chain: physics-built $W$ $\to$ spectral properties of $L$ $\to$ normalization $\tilde{L}=L/\lambda_{\max}$ $\to$ controlled spectral radius of $P$. Removing any link breaks the control; the ablations and spectral-radius measurements of Section~\ref{sec:4} (E3, Table~\ref{tab:4_7}) test this chain when normalization is removed.

MPNO is neither a method specialized to one PDE family nor a minor modification of existing architectures, but an architecture-level solution to autoregressive instability of neural operators; its spectral constraint relies only on nonnegative symmetry of $W$ and connectivity of the graph, independent of the specific PDE. Three scenarios of different types (hyperbolic shocks in Burgers, elliptic flow in Darcy, and concrete penetration) validate the method's autoregressive stability and the cross-scenario transferability of the edge-weight formula (\S\ref{sec:4}).

MPNO is a constitutive Markov neural-operator architecture, where ``constitutive'' means that the edge weights encode material-interface constitutive/impedance information through the acoustic-impedance harmonic mean and the traction amplitude (\S\ref{sec:3_4}), embedding mesoscale material behavior as a physical prior into the propagation structure. Its skeleton is the three-component chain physics-coupled adjacency $\to$ Markov propagation operator $\to$ Rayleigh-quotient spectral estimation, and it transfers to different PDEs by replacing the coupled material-property variables (acoustic impedance $\to$ permeability $\to$ uniform weights). The main contributions are:

(i) a Markov (row-stochastic) propagation-operator architecture that models one-step evolution as a linear propagation $P=I-\alpha\tilde{L}$ with controlled spectral radius, suppressing exponential amplification of autoregressive errors by mechanism; stability thus becomes a designable architectural property at the propagation-operator level rather than an optimized loss objective;

(ii) a physics-coupled edge-weight construction (acoustic-impedance harmonic mean $\times$ contact area $\times$ traction amplitude $\times$ zero-initialized MLP correction) that embeds material-interface constitutive/impedance information into graph propagation as a zero-parameter physical skeleton, and is transferable across PDEs by replacing material-property variables alone;

(iii) an online Rayleigh-quotient power iteration that engineers the spectral-radius constraint into the training loop (decoupled from the optimization trajectory), validated on Burgers, Darcy, and concrete-penetration PDEs for non-divergence, cross-scenario transferability, and lightness at about 20K parameters.

The lineage of neural operators traces to DeepONet \cite{ref6}, whose inner product of a branch network (encoding the input function) and a trunk network (encoding the output location) constitutes the first operator-level mapping framework; a systematic theory of operator learning by composition of integral operators is given in \cite{ref16}. Graph neural operators (GNO) \cite{ref7} learn integral kernels on graphs through message passing, yet tend to be unstable for deep networks; the edge-message-passing paradigm draws on MPNN \cite{ref18}, graph convolutional networks (GCN) \cite{ref19}, and graph attention networks (GAT) \cite{ref20}. The Fourier neural operator (FNO) \cite{ref8} replaces message passing with global Fourier basis functions for spectral-domain convolution, achieving high accuracy and efficiency on Burgers, Darcy, and Navier--Stokes equations, under the core assumption that solutions are dominated by low-order Fourier modes. WNO \cite{ref1} further substitutes wavelet bases for Fourier bases and, through joint time--frequency localization, outperforms FNO on problems with discontinuities and spikes; the recent U-shaped neural operator (U-NO) \cite{ref17} uses U-Net-style down/up-sampling for deeper, more memory-efficient operator learning. MeshGraphNets \cite{ref9} (formerly GNS \cite{ref21}) represent a mesh as a graph and learn per-edge message functions, advancing fluid and rigid-body dynamics, but their per-edge MLP message functions impose no constraint on the spectral radius of the effective propagation operator. Graph-diffusion methods such as SGC \cite{ref14} and APPNP \cite{ref15} spread information on graphs through normalized propagation or personalized PageRank, and their propagation operator $P$ likewise carries spectral properties; unlike these methods aimed at normalized graph convolution, MPNO constructs $P=I-\alpha\tilde{L}$ from physics-coupled edge weights and employs the spectral-radius constraint as an architectural guarantee of autoregressive stability for mechanical surrogate models (\S\ref{sec:3_2}--\S\ref{sec:3_5}).

Physics injection forms a separate lineage. PINNs \cite{ref2} add PDE residuals as loss penalties so that the network obeys physics even with sparse data; PINOs \cite{ref3} carry this idea into operator learning; a systematic survey of physics-informed machine learning is \cite{ref27}, with antecedents in hidden-physics models \cite{ref40} and domain-decomposition variants in XPINN \cite{ref28}. Both routes, however, impose physics as a soft constraint, making conservation and stability optimized loss objectives rather than algebraic architectural properties; their gradient ill-conditioning and discontinuity difficulties were noted above. Within the ``how physics is injected'' lineage, three approaches lie closer to the aforementioned mechanics setting: the peridynamic neural operator (PNO) \cite{ref41} hard-guarantees momentum conservation and objectivity in-architecture from a state-based peridynamic construction, with a purely data loss; constitutive artificial neural networks (CANN) \cite{ref42} embed thermodynamics (stress derived analytically from strain-energy density) and polyconvexity in the architecture, taking invariants of the right Cauchy--Green tensor $C=F^{\top}F$ to ensure objectivity and symmetry; and physics-informed MeshGraphNets (PI-MGNs) \cite{ref43} use FEM weak-form residuals as a label-free training loss. These exemplify three modes of physics injection: in-architecture conservation/objectivity, in-architecture thermodynamics/polyconvexity, and weak-form residual losses.

Two clarifications position MPNO relative to these approaches. First, regarding how physics is imposed: MPNO pursues a different mode of physics injection from these three: it neither hard-constrains the constitutive law in-architecture as CANN \cite{ref42} does (thermodynamics and polyconvexity, an idea from input-convex networks \cite{ref10}), nor hard-guarantees momentum conservation and objectivity in-architecture as PNO \cite{ref41} does, nor adopts weak-form residual losses as PI-MGNs \cite{ref43} do. Instead, material-interface constitutive information (acoustic impedance and traction) is encoded directly into the edge weights $W$ through a zero-parameter formulaic skeleton (\S\ref{sec:3_4}), a physically motivated inductive bias that adds no parameters and transfers across scenarios. Second, regarding how stability is guaranteed: MPNO superficially resembles regularization such as spectral normalization \cite{ref11}, but differs essentially: spectral normalization imposes Lipschitz constraints on weight matrices during training, and its effect depends on the optimization process. A number of alternative designs for the long-horizon stability of autoregressive neural operators have recently emerged around stability mechanisms: SGNO \cite{ref44} models the linear part of the propagation operator with a constrained real non-positive diagonal spectral generator, paired with a gated forcing term for the nonlinear part, and gives single-step amplification and finite-horizon rollout bounds that, however, depend on local Lipschitz constants with the nonlinear component left explicitly unconstrained; SpectraNet \cite{ref45} parameterizes a residual target $f_\theta=\mathrm{id}+\Delta_\theta$ with a semigroup-consistency loss, improving the exponential growth of the rollout error, $O(L^T\varepsilon_0)$, to linear drift $O(T\delta)$, but its guarantee relies on convergence of the training loss; Thermalizer \cite{ref46} targets time-stationary systems with an invariant measure, estimating the score of the invariant measure at inference with a diffusion model and denoising rollout states frame by frame to extend the stable prediction horizon. These three achieve stability through network spectral structure, training loss, and inference-time correction, respectively, still training/inference-level guarantees. MPNO's spectral constraint ($\rho(P)\le1$) is realized by estimating $\lambda_{\max}$ online via power iteration and normalizing, guaranteed by the Markov propagation operator built from physics-coupled edge weights; it is independent of training and requires no inference-time correction. MPNO's differentiating combination is the coupling of constitutive-embedded edge weights with constructive spectral stability, rather than any single physics hard constraint.

These four lineages (operator learning, graph methods, physics-informed learning, and stability regularization) each correspond to one facet of MPNO. The spectral-radius constraint itself is classical in linear stability theory and spectral graph theory \cite{ref22}: the spectral properties of normalized Laplacians \cite{ref23}, PageRank as Markov random walks \cite{ref24}, graph-diffusion kernels \cite{ref25}, and label propagation \cite{ref26} have all established the spectral structure of graph propagation operators $P$. No prior work has built a Markov propagation operator from a physics-coupled graph Laplacian and engineered the spectral-radius constraint into a neural-operator training loop via online power iteration; MPNO's contribution is the engineered realization of this combination and its validation in mechanics scenarios, not a new spectral criterion.

In what follows, Section~\ref{sec:2} defines the problem setting and data generation; Section~\ref{sec:3} develops the full mathematical framework of MPNO; Section~\ref{sec:4} validates the core contributions through three-scenario experiments and ablations; Section~\ref{sec:4_5} discusses applicability and limitations; Section~\ref{sec:5} concludes.

\section{Problem Setting and Data Generation}
\label{sec:2}

This section defines the three PDE problems and the construction of their datasets, covering the three mathematical types needed to validate MPNO. A systematic treatment of data-driven scientific and engineering modeling (including PDE surrogates and system identification) is given in \cite{ref39}. Data for the Burgers and Darcy equations are generated directly from analytic solutions or high-accuracy numerical solvers. Concrete-penetration data, generated by LS-DYNA explicit-dynamics simulation, serve as the extreme engineering validation scenario. Beyond these three trained scenarios, Section~\ref{sec:repro_euler} reports an out-of-distribution generalization check on the two-dimensional compressible Euler equations (generated by the PDEBench~\cite{ref30} pipeline), verifying that the stability guarantee transfers to genuinely transient flow outside the training distribution.

\subsection{Burgers equation (1D hyperbolic PDE)}
\label{sec:2_1}

The governing equation is $\partial_t u + u\partial_x u = \nu \partial_x^2 u$, $x \in [0,1]$, with periodic boundary conditions. Analytic solutions for arbitrary initial conditions follow from the Cole--Hopf transform. 100 initial conditions are generated as random superpositions of sine waves ($N = 128$ uniform grid), $\nu = 0.01$, with $t \in [0,1]$ discretized into 50 steps; 30 are used for training and the remaining 70 for testing, with 5 rollout-stability seeds drawn from the test set.

\subsection{Darcy flow (2D elliptic PDE)}
\label{sec:2_2}

The governing equation is $-\nabla \cdot (K(\mathbf{x}) \nabla p(\mathbf{x})) = f(\mathbf{x})$, $\mathbf{x} \in [0,1]^2$, with Dirichlet boundary conditions. The permeability field $K(\mathbf{x}) = \exp(g(\mathbf{x}))$ is a Gaussian random field generated by FFT (Mat\'ern covariance, $\tau = 3$) with values in $[0.1, 10]$, spanning the full physical spectrum from low to high permeability. The pressure field is solved by a five-point finite-difference scheme (grid $64 \times 64$, $N = 4096$). The training set comprises 100 random $K(\mathbf{x})$ fields and the test set 20.

\subsection{Concrete penetration (3D transient dynamics)}
\label{sec:2_3}

Depth prediction for concrete penetration has mature empirical formulas \cite{ref36} and perforation-test benchmarks \cite{ref37}; common macroscopic constitutive models include HJC \cite{ref34}, RHT \cite{ref35}, and CSCM. The data-generation pipeline follows the random-aggregate placement (RSA) framework of Wu et al.\ \cite{ref12}. In a $\phi 500 \times 200$ mm cylindrical domain, a 3D spherical (crushed) granite aggregate distribution with 42\% volume fraction and 5--25 mm sizes is generated by Fuller gradation ($n = 0.5$), with KD-Tree-accelerated collision detection. Penetration simulations run in LS-DYNA/Explicit \cite{ref38}: a $\phi 45$ mm ogive-nose steel projectile (0.8 kg) impacts normally at four velocities (100/135/165/200 m/s), with 100 aggregate distributions per velocity; half of the cases use a pure ogive nose and the other half a chamfered-ogive nose, which yields nearly identical penetration response with a higher simulation success rate. The three lower velocities (100/135/165 m/s) form the evaluation set; the 200 m/s case participates in training only as diversity-augmentation data and is excluded from evaluation (\S\ref{sec:3_7_1}): at high impact velocity the inter-frame stress evolution is relatively slow and adjacent frames are highly similar, offering low discrimination as a single-step evaluation baseline, yet its samples enrich the training distribution with higher-amplitude, steeper-gradient regimes (Table \ref{tab:4_7}). Aggregates use the JH-2 constitutive model \cite{ref13} and the mortar the CSCM model \cite{ref33}; the solution time is 700 $\mu$s with a 17.5 $\mu$s step, outputting 40 frames. Stress fields are trilinearly interpolated onto a $64 \times 64$ Cartesian grid and normalized by $f_c = 30$ MPa; training and evaluation are performed on an impact-zone subgraph (about 200 nodes, varying between 165 and 238 across seeds, with an active-node ratio of about 30\%) to focus on the stress-wave-dominated local region and mitigate global sparsity. The aggregate volume-fraction field $V_f \in [0,1]^{64 \times 64}$ is computed by sub-pixel sampling as a continuous encoding of mesoscale heterogeneity. The ITZ ($\sim$50 $\mu$m) is homogenized into the mortar phase. The dataset comprises 400 independent effective samples (100 per velocity), split 70:15:15 into training/validation/test sets. Each sample contains a stress time series $\mathbf{S} \in \mathbb{R}^{6 \times 64 \times 64 \times 40}$ and a volume-fraction field $V_f \in \mathbb{R}^{64 \times 64}$. Table \ref{tab:1} summarizes the key parameters.

\begin{table}[!ht]
	\centering
	\caption{Parameters of the concrete-penetration dataset}
	\label{tab:1}
	\begin{tabularx}{\textwidth}{l X}
		\toprule
		Parameter & Value \\
		\midrule
		Target & $\phi 500 \times 200$ mm (cylinder) \\
		\addlinespace
		Aggregate fraction / size & 42\% / 5--25 mm \\
		\addlinespace
		Aggregate / mortar constitutive & JH-2 (granite) / CSCM (C30) \\
		\addlinespace
		Projectile / velocity & $\phi 45$ mm ogive steel projectile / 100, 135, 165, 200 m/s (100 each) \\
		\addlinespace
		Grid resolution / time frames & $64 \times 64$ / 40 frames \\
		\addlinespace
		ITZ & Homogenized dispersion ($\sim$50 $\mu$m) \\
		\bottomrule
	\end{tabularx}
\end{table}

\begin{figure}[!ht]
	\centering
	\includegraphics[width=\linewidth]{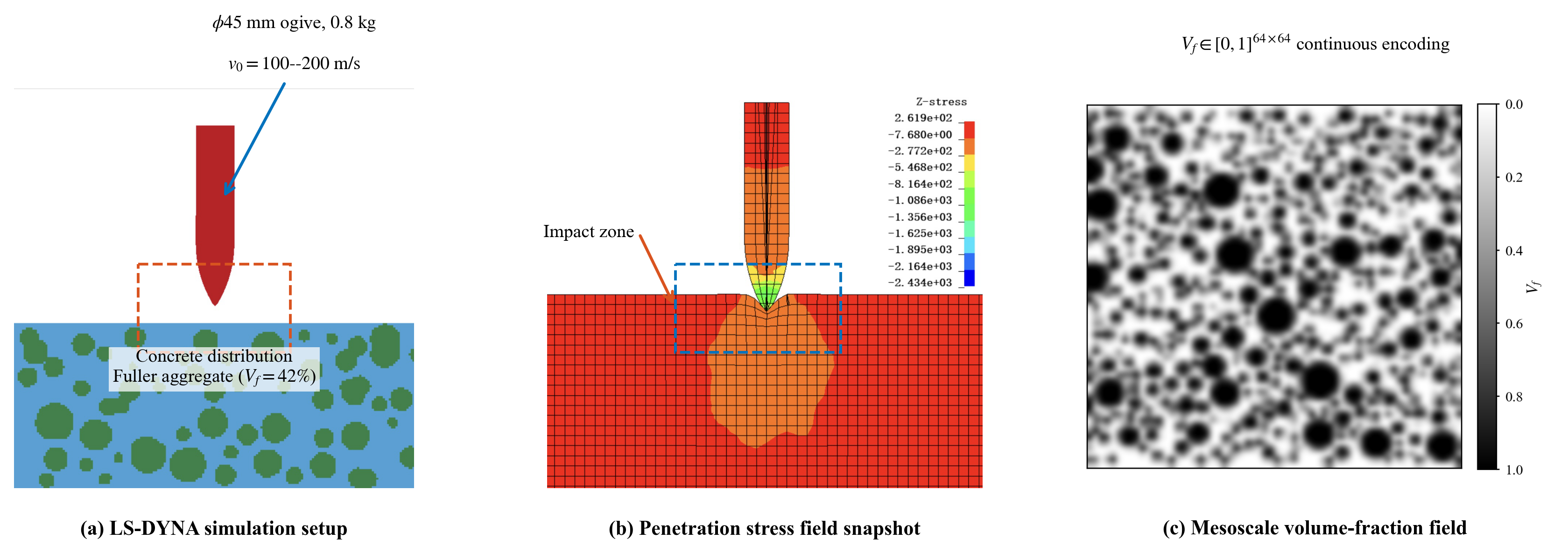}
	\caption{Problem geometry and data. (a) LS-DYNA setup: a $\phi$500$\times$200 mm cylindrical concrete target (Fuller-graded aggregates 5--25 mm, 42\% volume fraction, JH-2/CSCM) impacted by a $\phi$45 mm ogive-nose steel projectile (0.8 kg, 100--200 m/s). (b) An LS-DYNA $\sigma_{zz}$ stress-field snapshot of the impact zone (seed 5, 100 m/s, frame 20; normalized by 30 MPa, range $\pm 8.5$; peak $\sigma_{zz}$ 5.039). (c) An example aggregate volume-fraction field $V_f$.}
	\label{fig:2}
\end{figure}

Fig.~\ref{fig:2} illustrates the problem geometry, an LS-DYNA stress field, and the aggregate volume-fraction field.

\section{The Constitutive Markov Physics-Informed Neural Operator (MPNO)}
\label{sec:3}

\subsection{Problem formulation: from continuum to discrete graph}
\label{sec:3_1}

MPNO departs from the conventional neural-operator paradigm that models transient dynamics as a map between continuous function spaces, and instead formulates it as a Markov process on a spatially discretized graph. This section develops the mathematical framework of this reformulation: defining the graph topology, mapping the stress field to node-state vectors, and giving the basic construction of the adjacency matrix and the propagation operator.

\begin{quote}
\textbf{Evaluation-configuration statement.} The mathematical framework developed in \S\ref{sec:3_1}--\S\ref{sec:3_8} is defined in general form, and its spectral conclusions (Property 1 and the spectral analysis) hold for both the 3D hexahedral-mesh configuration (6-face adjacency, Cauchy traction decomposition, three-frame delayed states, $E \approx 3N$) and the 2D transverse-section configuration (adjacency $\pm x, \pm y$, with $z$ the out-of-plane penetration direction) (dependency conditions in \S\ref{sec:3_1_4}). The numerical experiments instantiate and validate the \textbf{2D transverse-section configuration}: nodes lie in the $x$--$y$ plane ($z$ is the out-of-plane penetration direction with coordinate identically zero), node features are single-frame 10-dimensional vectors $[\boldsymbol{\sigma}, \mathbf{x}_i, v_i] \in \mathbb{R}^{10}$ (6 stress components including the out-of-plane $\sigma_{zz}$, 3 normalized spatial coordinates, 1 velocity), adjacency is 4-directional ($\pm x, \pm y$), and $E \approx 2N$ edges. The end-to-end data flow of the evaluation configuration is summarized in \S\ref{sec:3_11}.
\end{quote}

\subsubsection{Mesh-to-graph topology mapping}
\label{sec:3_1_1}

Consider a spatial domain discretized into a hexahedral mesh of $X \times Y \times Z$ cells. Define a graph $\mathcal{G} = (\mathcal{V}, \mathcal{E}, W)$, where:

\begin{itemize}
\item \textbf{node set} $\mathcal{V} = \{1, 2, \dots, N\}$, $N = X \times Y \times Z$, one node per cell
\item \textbf{edge set} $\mathcal{E} \subseteq \mathcal{V} \times \mathcal{V}$, with $(i, j) \in \mathcal{E}$ iff cells $i$ and $j$ share a contact face
\item \textbf{adjacency weight matrix} $W = [w_{ij}] \in \mathbb{R}^{N \times N}_{\ge 0}$, with $w_{ij} > 0$ when $(i,j) \in \mathcal{E}$ and $w_{ij} = 0$ otherwise
\end{itemize}

Interior cells have exactly 6 neighbors (along $\pm x, \pm y, \pm z$); boundary cells have fewer. The graph has $E = |\mathcal{E}| \approx 3N$ edges (each interior edge is shared by two directed edges in the undirected graph).

Six-face adjacency is not a freely chosen hyperparameter: structured hexahedral meshes are the standard discretization of explicit-dynamics solvers, and each interior cell interacts with its neighbors through exactly 6 faces. The Cauchy stress tensor $\boldsymbol{\sigma}\in\mathbb{R}^{3\times3}_{\text{sym}}$ has 6 independent components, matching the 6 face-normal directions, and the tractions on the 6 faces jointly constitute a complete representation of $\boldsymbol{\sigma}$ (fewer faces underdetermine it; more faces add redundancy bounded by its 6 degrees of freedom). Six faces are thus dictated jointly by physics and the mesh, not chosen as a free architecture parameter. Generalization to tetrahedral or unstructured meshes is discussed in \S\ref{sec:3_1_4}.

From the adjacency weight matrix, two derived matrices can be defined:

\[
D = \mathrm{diag}(d_1, d_2, \dots, d_N), \quad d_i = \sum_{j=1}^N w_{ij} \quad \text{(degree matrix)}
\]

\[
L = D - W \quad \text{(unnormalized graph Laplacian)}
\]

\begin{figure}[!ht]
	\centering
	\includegraphics[width=\linewidth]{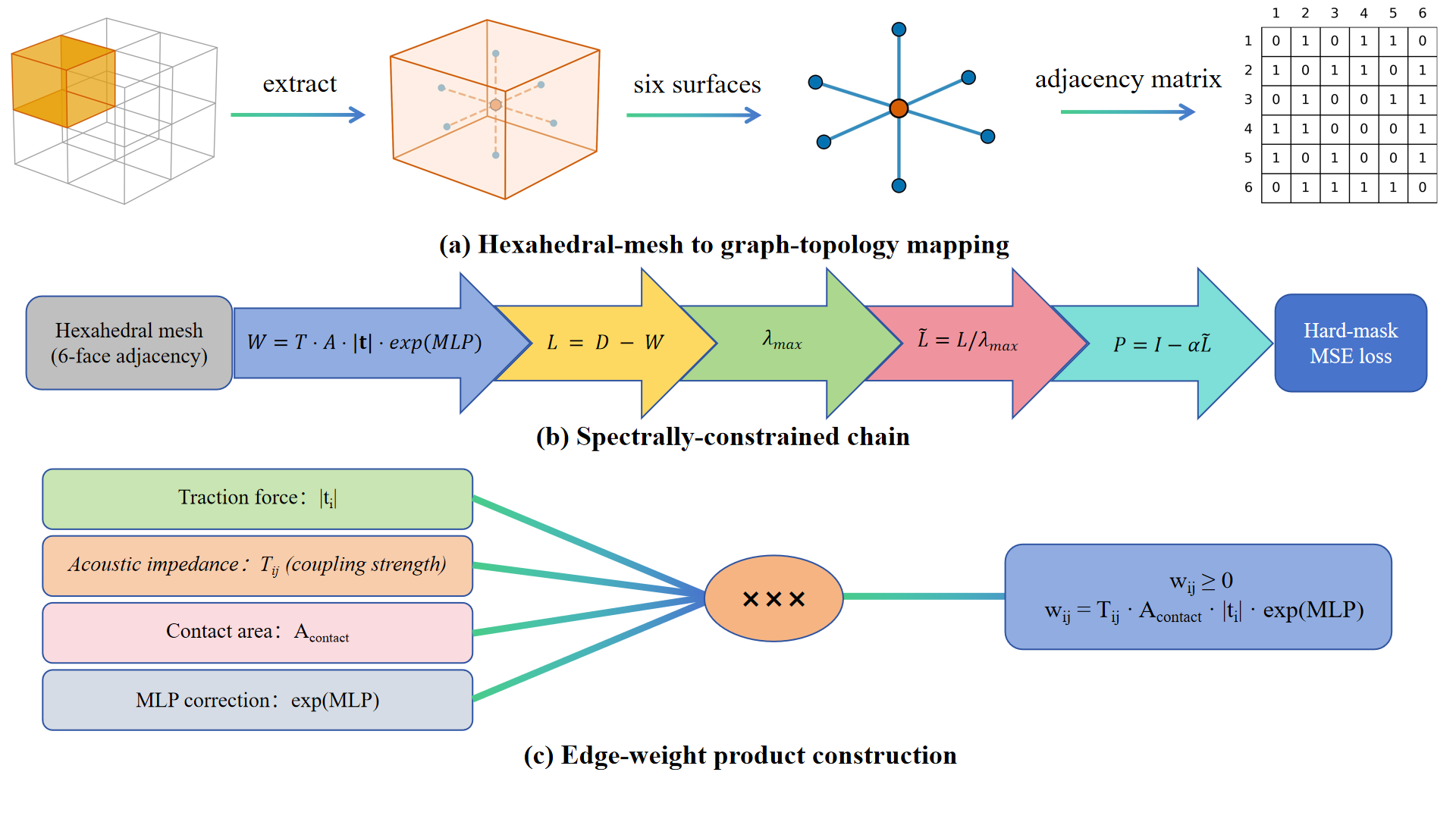}
	\caption{MPNO method. (a) Hexahedral-mesh to graph-topology mapping (2-D section schematic; the 3-D framework uses 6-face adjacency, $E\approx3N$). (b) Spectrally-constrained chain: hexahedral mesh $\to$ adjacency $W=T\cdot A\cdot|\mathbf{t}|\cdot\exp(\mathrm{MLP})$ (nonneg.\ symmetric, \S\ref{sec:3_4}) $\to$ Laplacian $L=D-W$ (PSD, \S\ref{sec:3_2}) $\to$ Rayleigh-quotient power iteration $\hat{\lambda}_{\max}$ ($O(E)$ online estimate, \S\ref{sec:3_2_2}) $\to$ normalized $\tilde{L}=L/\lambda_{\max}$ (spectrum in $[0,1]$) $\to$ Markov propagator $P=I-\alpha\tilde{L}$ ($\rho(P)=1$, Property 1) $\to$ decoder $\to$ hard-masked MSE loss. Unlike the one-sided Rayleigh-quotient estimate (\S\ref{sec:3_2_2}), the spectral constraint $\rho(P)\le1$ is closed and exact; this closed constraint is the mechanism underlying autoregressive stability. (c) Edge-weight product construction: $T_{ij}$ (acoustic impedance, harmonic mean $2Z_iZ_j/(Z_i+Z_j)$) $\cdot$ $A_{\mathrm{contact}}$ (contact area, fixed $=1$) $\cdot$ $|\mathbf{t}_i|$ (traction magnitude) $\cdot$ $\exp(\mathrm{MLP})$ (zero-init correction) $\to w_{ij}\ge0$ (nonneg., symmetric).}
	\label{fig:3}
\end{figure}

The diagonal entries $d_i$ of $D$ measure the total coupling strength incident on node $i$; $L$ is the second-order difference operator of the graph, whose properties (symmetric positive semidefinite) are analyzed in \S\ref{sec:3_2}. Fig.~\ref{fig:3} illustrates the hexahedral-mesh-to-graph mapping.

\subsubsection{Cauchy traction decomposition and node states}
\label{sec:3_1_2}

Each cell center stores a Cauchy stress tensor (6 components in Voigt notation). In the 3D framework, face tractions are computed via the Cauchy formula $\mathbf{t} = \boldsymbol{\sigma}\cdot\mathbf{n}$; the $6\ \text{faces} \times 3$ components form an 18-dimensional traction vector $\mathbf{t}_i \in \mathbb{R}^{18}$, and the map $\mathbb{R}^6 \to \mathbb{R}^{18}$ is invertible (the 6-dimensional stress tensor is recovered by linearly combining the positive/negative faces). Introducing three-frame delays and coordinates yields the augmented state $\mathbf{h}_i^t = [\mathbf{t}_i^t, \mathbf{t}_i^{t-1}, \mathbf{t}_i^{t-2}, \mathbf{x}_i] \in \mathbb{R}^{56}$, whose differences encode strain-rate and acceleration information. The evaluation configuration uses single-frame 10-dimensional node features (as stated in \S\ref{sec:3_1}).

\subsubsection{Edge orientation and face pairing}
\label{sec:3_1_3}

Each edge $(i,j)$ implies a face-to-face relationship: a face of node $i$ contacts the reverse face of node $j$, encoded as a face-pairing index $(\phi_i, \phi_j)$. Face-pairing information is indispensable to the physics-coupled edge weights, because the traction-amplitude coupling must know on which face the force acts.

\subsubsection{Generality of the framework}
\label{sec:3_1_4}

The mathematical framework of MPNO (Property 1 and the spectral analysis) depends only on nonnegative symmetry of $W$ and connectivity of the graph, not on node degrees or the edge-set definition; its spectral conclusions therefore extend to tetrahedral and polyhedral meshes (graph topology or weight interpolation must be adapted, but this property provides an invariant mathematical foundation).

\subsection{Graph-Laplacian propagation and spectral stability}
\label{sec:3_2}

\S\ref{sec:3_1} expressed the impact-dynamics system as a state-evolution problem on the graph $\mathcal{G} = (\mathcal{V}, \mathcal{E}, W)$. This section establishes the core mathematical mechanism of this evolution: constructing the propagation matrix $P$ via the graph Laplacian, and showing that its spectral radius can be estimated online by Rayleigh-quotient power iteration for $\lambda_{\max}$ and constrained by normalization to $\rho(P) \le 1$.

\subsubsection{Spectral properties of the graph Laplacian}
\label{sec:3_2_1}

Starting from the nonnegative symmetric adjacency matrix $W$ and the degree matrix $D$, the unnormalized graph Laplacian is constructed:

\[
L = D - W \in \mathbb{R}^{N \times N}
\]

By the nonnegative symmetry of $W$, $L$ is real symmetric positive semidefinite: its quadratic form $\mathbf{x}^T L \mathbf{x} = \sum_{(i,j)\in\mathcal{E}} w_{ij}(x_i-x_j)^2 \ge 0$ measures the smoothness of the state vector on the graph, with equality iff the state is uniform on every connected node pair, giving the zero eigenvalue $\lambda_1=0$ with eigenvector $\mathbf{1}$ (the uniform mode is not attenuated).

The normalized Laplacian $\tilde{L} = L / \lambda_{\max}(L)$ is further defined, where $\lambda_{\max}$ is the spectral radius of $L$. Scalar scaling does not change the eigenvector basis; from $L\mathbf{v}_i = \lambda_i \mathbf{v}_i$ it follows that $\tilde{L}\mathbf{v}_i = (\lambda_i / \lambda_{\max}) \mathbf{v}_i$, so its eigenvalues $\{\tilde{\lambda}_i\}_{i=1}^N$ lie strictly in $[0,1]$: $0 = \tilde{\lambda}_1 \le \tilde{\lambda}_2 \le \dots \le \tilde{\lambda}_N = 1$.

\subsubsection{Online spectral-radius estimation: Rayleigh-quotient power iteration}
\label{sec:3_2_2}

Normalization requires $\lambda_{\max}(L)$, but full eigendecomposition at $O(N^3)$ cost is infeasible inside the training loop. Rayleigh-quotient power iteration is instead used, a lightweight spectral estimator requiring only matrix--vector products, which approximates $\lambda_{\max}$ online at $O(E)$ cost per training batch.

Power iteration starts from a random initial vector $\mathbf{b}_0 \in \mathbb{R}^N$ (orthogonal to $\mathbf{1}$ to avoid capturing the zero eigenvalue) and iterates:

\[
\mathbf{b}_{k+1} = \frac{L \mathbf{b}_k}{\|L \mathbf{b}_k\|_2}, \quad k = 0, 1, \dots, K-1
\]

After $K$ iterations, the Rayleigh quotient gives the estimate of $\lambda_{\max}$:

\[
\hat{\lambda}_{\max} = R(\mathbf{b}_K) = \frac{\mathbf{b}_K^T L \mathbf{b}_K}{\mathbf{b}_K^T \mathbf{b}_K}
\]

\textbf{Convergence.} $\hat{\lambda}_{\max}$ converges to $\lambda_{\max}$ at rate $O\left((\lambda_{N-1}/\lambda_{\max})^{2K}\right)$ (derivation in \S\ref{sec:3_5_2}); in practice $K=20$ brings the relative error to about $5\times10^{-6}$ (an underestimate of $0.0005\%$), and the training loop uses a smaller $K$ for efficiency.

In practice, an exponential moving average (EMA, decay $\rho = 0.99$) is applied across batches to $\hat{\lambda}_{\max}$:

\[
\bar{\lambda}_{\max}^{(t)} = \rho \cdot \bar{\lambda}_{\max}^{(t-1)} + (1-\rho) \cdot \hat{\lambda}_{\max}^{(t)}
\]

EMA suppresses batch-to-batch fluctuation of the spectral estimate caused by nonuniformity in the material-parameter space (the graph topology is fixed during training, so the cross-batch variation of $\lambda_{\max}$ is pure sampling noise), ensuring a smooth update of $\alpha$ (the propagation rate, defined below).

\subsubsection{Construction and stability of the propagation operator}
\label{sec:3_2_3}

Given the normalized Laplacian $\tilde{L}$, the Markov propagation operator is constructed:

\[
P = I - \alpha \tilde{L}, \quad \alpha = \sigma(\theta) \in [0, 1]
\]

where $\sigma(\cdot)$ is the sigmoid function and $\theta$ a learnable logit parameter. The propagation rate $\alpha$ controls the magnitude of the single-step state update: $\alpha = 0$ gives $P = I$ (no propagation, state unchanged) and $\alpha = 1$ gives maximal propagation.

\begin{proposition}[Property 1 (bounded spectral radius of Markov propagation)]
Let $P = I - \alpha \tilde{L}$ with $\alpha \in [0,1]$ and $\tilde{L}$ the normalized graph Laplacian (eigenvalues $\tilde{\lambda}_i \in [0,1]$). Then $P$ has eigenvalues $\mu_i = 1 - \alpha \tilde{\lambda}_i \in [0,1]$, so $P$ is real symmetric positive semidefinite with spectral radius $\rho(P) = \|P\|_2 = 1$; for any initial state $\mathbf{v}_0 \in \mathbb{R}^N$ and any rollout step $t \in \mathbb{N}^+$, the autoregressive sequence $\mathbf{v}_t = P^t \mathbf{v}_0$ satisfies the nonexpansive bound $\|\mathbf{v}_t\|_2 \le \|\mathbf{v}_0\|_2$. These properties are carried by the construction of $P$ ($\tilde{L}$ normalized via power iteration), not optimized as a loss objective.
\end{proposition}

\begin{quote}
\textbf{Implementation note.} The statement $\rho(P)=1$ presumes normalization of $\tilde{L}$ by the exact value of $\lambda_{\max}$. In practice, $\lambda_{\max}$ is estimated online by Rayleigh-quotient power iteration (only under-estimating, never over-estimating; measured mean underestimate $0.0005\%$, \S\ref{sec:4_2}), so the realized constraint is $\rho(P)\le1$, with the actual margin coming from the underestimation error (see Remark 1).
\end{quote}

Note that $W$ depends on the current-frame stress level ($|\mathbf{t}_i|$ and the stress magnitude/velocity in the MLP inputs, see \S\ref{sec:3_4_1}), so $W$ and $P$ are updated per frame (denoted $W^{(t)}$, $P_t$). Every per-frame $W^{(t)}$ is nonnegative symmetric, so the spectral properties and Property 1 hold frame by frame; moreover $\|P_t\|_2 = 1$ for each frame, so nonexpansiveness propagates along the time-varying chain $\mathbf{v}_{t+1} = P_t\mathbf{v}_t$. The state-dependent transition of $P$ depends only on the current state, so the Markov property is preserved; the long-horizon behavior of the full model is confirmed empirically in \S\ref{sec:4} (Remark 2, \S\ref{sec:4_5_2}).

\begin{remark}[Remark 1 (spectral constraint vs.\ a posteriori soft constraint)]
The core value of Property 1 is that the spectral-radius constraint is carried by the construction of $P = I - \alpha\tilde{L}$, rather than ``hoped for'' as a loss penalty during optimization: the former is realized by normalizing against the online power-iteration estimate of $\lambda_{\max}$, the latter depends on the chance of the training trajectory. The Fourier/wavelet convolution layers of FNO and WNO impose no structural constraint on the spectral radius of their effective propagation operator; whatever value $\rho(J_{f_\theta})$ takes during training is entirely decided by the chance of random initialization and the SGD trajectory. \S\ref{sec:3_3} will argue that, without a constrained spectral radius, long-horizon autoregressive stability admits no structural guarantee (stability depends on contraction emergent from data and training trajectory, and cannot be relied upon). ``Constraint'' here means: power iteration yields an estimate of $\lambda_{\max}$ (only under-estimating, never over-estimating), and normalizing by it makes $\rho(P)\le1$ hold in practice, with the margin coming from the underestimation error ($0.0005\%$); it does not promise an absolute guarantee beyond physics.
\end{remark}

\subsubsection{Iterative scheme of graph propagation}
\label{sec:3_2_4}

State vectors are propagated over the graph by $P$ in multiple rounds. The $r$-th round ($r = 1,\dots,R$, $R=8$) takes the form:

\[
\mathbf{v}^{(r)} = P \mathbf{v}^{(r-1)} = (I - \alpha \tilde{L}) \mathbf{v}^{(r-1)}
\]

Expanded into matrix--vector operations:

\[
\tilde{L} \mathbf{v} = \frac{1}{\bar{\lambda}_{\max}} (D\mathbf{v} - W\mathbf{v})
\]

where $D\mathbf{v}$ weights by node degree (elementwise product) and $W\mathbf{v}$ is realized by a sparse matrix--vector product in which each edge $(i,j)$ aggregates $\mathbf{v}_i \cdot w_{ij}$ to the target node $j$. The $O(E)$ cost of this operation keeps multi-round propagation affordable: $R=8$ rounds total $O(8 \cdot E \cdot d_{\text{hidden}})$, about 45\% of a single-step inference (see the complexity analysis in \S\ref{sec:3_9}).

After each propagation round, an APPNP-style initial residual connection ($\beta = 0.3$) is applied to prevent over-smoothing:

\[
\mathbf{v}^{(r)} = (1-\beta) \cdot P\mathbf{v}^{(r-1)} + \beta \cdot \mathbf{v}^{(0)} + \text{MLP}_r(\mathbf{v}^{(r)})
\]

where $\mathbf{v}^{(0)}$ is the initial node embedding before propagation (encoder output) and $\text{MLP}_r$ is the two-layer perceptron of round $r$ (with LayerNorm). The residual connection ensures that node representations retain their initial local features even after 8 propagation rounds, which is necessary for simultaneously capturing long-range wave propagation and local material scattering, the multiscale features pervasive in transient dynamics.

\subsubsection{Three-component chain and the hard-constraint boundary}
\label{sec:3_2_5}

The three components (the physics-coupled adjacency matrix $W$ (\S\ref{sec:3_4}, nonnegative symmetric), the Markov propagation operator $P$, and the Rayleigh-quotient power iteration) form a one-way dependency chain that draws the boundary between MPNO and methods imposing physics through loss functions: \textbf{stability ($\rho(P)\le1$) is constructed into the structure of $P$ as an architectural property, not as a loss penalty}, while the physical prior (constitutive impedance) is encoded into the edge weights $W$ as an inductive bias on a different link of the chain. Neither enters the loss or generates a physical-residual gradient, and neither relies on the training trajectory (Property 1). \S\ref{sec:3_3} argues the consequences of severing either link.

\subsection{Spectral-radius criterion for autoregressive collapse}
\label{sec:3_3}

Property 1 showed that the spectral radius of MPNO's propagation operator $P = I - \alpha\tilde{L}$ can be estimated via power iteration for $\lambda_{\max}$ and constrained by normalization to $\rho(P) \le 1$, bounding the error growth of autoregressive rollout at the propagation-operator level. This section argues the converse: without a constrained spectral radius, autoregressive rollout destabilizes. A general mathematical criterion is first established: the long-term behavior of any linear or locally linearized autoregressive system is strictly dichotomized by the spectral radius of its propagation matrix, and then apply it to explain the observations on the penetration data (autoregressive instability of WNO/FNO; numerical values in \S\ref{sec:3_3_3}).

\subsubsection{Problem setup: from teacher forcing to autoregressive rollout}
\label{sec:3_3_1}

During training, neural operators commonly use teacher forcing: the model receives the true stress field $\sigma_t^{\text{GT}}$ as input and predicts $\sigma_{t+1}$, with the loss measuring the single-step deviation between prediction and ground truth:

\[
\mathcal{L}_{\text{train}} = \frac{1}{T} \sum_{t=0}^{T-1} \|f_\theta(\sigma_t^{\text{GT}}) - \sigma_{t+1}^{\text{GT}}\|_2^2
\]

Under this paradigm the model need only learn a single-step map from the ground-truth manifold to itself, and prediction errors are not accumulated.

At inference, the model switches to autoregressive rollout:

\[
\hat{\sigma}_{t+1} = f_\theta(\hat{\sigma}_t), \quad \hat{\sigma}_0 = \sigma_0^{\text{GT}}
\]

The prediction $\hat{\sigma}_t$ is fed back to generate $\hat{\sigma}_{t+1}$: the initial error $\epsilon_0 = \hat{\sigma}_1 - \sigma_1^{\text{GT}}$ becomes part of the next input and compounds through the nonlinear dynamics $f_\theta$. This training--inference distribution shift, known in the machine-learning literature as exposure bias, is mitigated by pushforward training with temporal bundling \cite{ref29} and iterative refinement \cite{ref31}; benchmarks such as PDEBench \cite{ref30} characterize autoregressive accuracy and stability, and autoregressive rollout can be viewed as the explicit discretization of a neural ODE \cite{ref32}.

\subsubsection{Linearization analysis: spectral radius as the collapse criterion}
\label{sec:3_3_2}

Take a first-order Taylor expansion of $f_\theta$ about the ground-truth trajectory $\sigma_t^{\text{GT}}$:

\[
\hat{\sigma}_{t+1} \approx f_\theta(\sigma_t^{\text{GT}}) + J_t \cdot (\hat{\sigma}_t - \sigma_t^{\text{GT}})
\]

where $J_t = \partial f_\theta / \partial \sigma \big|_{\sigma_t^{\text{GT}}} \in \mathbb{R}^{d \times d}$ is the Jacobian of $f_\theta$ at time $t$. Setting the error $\mathbf{e}_t = \hat{\sigma}_t - \sigma_t^{\text{GT}}$ and neglecting the single-step prediction residual $f_\theta(\sigma_t^{\text{GT}}) - \sigma_{t+1}^{\text{GT}}$ (assuming sufficient training), the error dynamics reduce to a linear system:

\[
\mathbf{e}_{t+1} \approx J_t \cdot \mathbf{e}_t
\]

Over $T$ autoregressive steps, the accumulated error depends on the Jacobian product $\prod_{t=0}^{T-1} J_t$. To obtain a general criterion, consider the effective average propagation matrix $P$. The long-term behavior of $\mathbf{e}_{t+1} = P \mathbf{e}_t$ is strictly dichotomized by the spectral radius $\rho(P) = \max_i |\mu_i(P)|$: if $\rho(P) < 1$, then $\lim_{t\to\infty}\|\mathbf{e}_t\|_2 = 0$ for any $\mathbf{e}_0 \in \mathbb{R}^d$ (the error converges to zero and the prediction tends to a fixed point of the propagation operator); if $\rho(P) > 1$, then for almost every $\mathbf{e}_0 \in \mathbb{R}^d \setminus \mathcal{S}$ ($\mathcal{S}$ a proper subspace of $\mathbb{R}^d$), $\lim_{t\to\infty}\|\mathbf{e}_t\|_2 = \infty$ (catastrophic divergence). The dichotomy follows from the Jordan normal form of $P$ (exponential decay of all blocks dominates polynomial growth when $\rho<1$; components along the dominant direction amplify when $\rho>1$).

The controlled regime is therefore $\rho(P)\le1$: if the spectral radius departs from 1 uncontrolled, downward (dissipation) or upward (amplification), autoregressive rollout inevitably fails on a sufficiently long time scale. Emergent training may reach this regime by chance, but a dependable guarantee requires the constraint to be carried by the architecture.

\subsubsection{Collapse of FNO/WNO: corollary and analysis}
\label{sec:3_3_3}

Neural operators such as FNO and the WNO developed earlier impose no structural constraint on the spectral radius of their Jacobians through their Fourier/wavelet convolution layers. Under teacher-forced training the model only minimizes the single-step prediction loss, so the spectral properties of the implicit Jacobian $J_{f_\theta}$ are entirely decided by the chance of random initialization and the SGD trajectory. In autoregressive rollout, if $\rho(J_{f_\theta}) < 1$ the model over-dissipates: high-frequency physical features decay exponentially in time and the predicted field tends to a contraction fixed point (nonzero under bias, so the field smooths rather than vanishes); if $\rho(J_{f_\theta}) > 1$, truncation noise and stepwise errors are exponentially amplified and the stress field diverges to numerical infinity. Hence, as long as the architecture lacks an a priori spectral-radius constraint, long-horizon autoregressive stability admits no structural guarantee (stability can rely only on contraction emergent from random initialization and the training trajectory, not on a designable, dependable architectural property); single-step accuracy under teacher forcing, however high, offers no guarantee of autoregressive stability.

\begin{remark}[Remark 2 (critical stability of MPNO and experimental evidence)]
Combining Property 1 with the spectral-radius dichotomy, MPNO's propagation operator $P = I - \alpha\tilde{L}$ satisfies $\rho(P) = 1$ exactly, placing it precisely on the critical boundary between dissipation and divergence. This is not a coincidence: the largest eigenvalue $\mu_{\max} = 1$ (with associated eigenvector $\mathbf{1}$) corresponds to the translation-invariant mode of global conservation. Note that this spectral property characterizes critical stability at the propagation-operator level; the full encode--propagate--decode model, owing to the nonlinear map and the energy relaxing toward the ground truth, may exhibit a measured energy-retention ratio below 1 (defined as the relative $L_2$-norm ratio $\|\hat{\mathbf{u}}_t\|_2/\|\mathbf{u}_0\|_2$ between the predicted and initial fields, see \S\ref{sec:4_1_1}; e.g., Burgers retains an energy ratio of 0.21 at 30 steps). This is model-level dissipation, not operator-level collapse; the two are reconciled by distinguishing $\rho(P)=1$ from the energy-decay rate.
\end{remark}

The predictions of the dichotomy agree with the experiments (full results in \S\ref{sec:4_3}). On the 400-sample penetration data, WNO ($\approx$11K parameters) reaches a single-step Rel-L2 of 0.7422 yet its autoregressive rollout diverges on all seeds (Rel-L2 at the $10^{12}$ scale); FNO ($\approx$79K parameters) reaches 0.7210 and stays stable on clean test seeds, but without any structural spectral-radius guarantee: the Fourier/wavelet kernels impose no constraint on $\rho(J_{f_\theta})$, and the departure direction of $\rho$ from 1 decides whether rollout diverges or decays.

\begin{figure}[!ht]
	\centering
	\includegraphics[width=\linewidth]{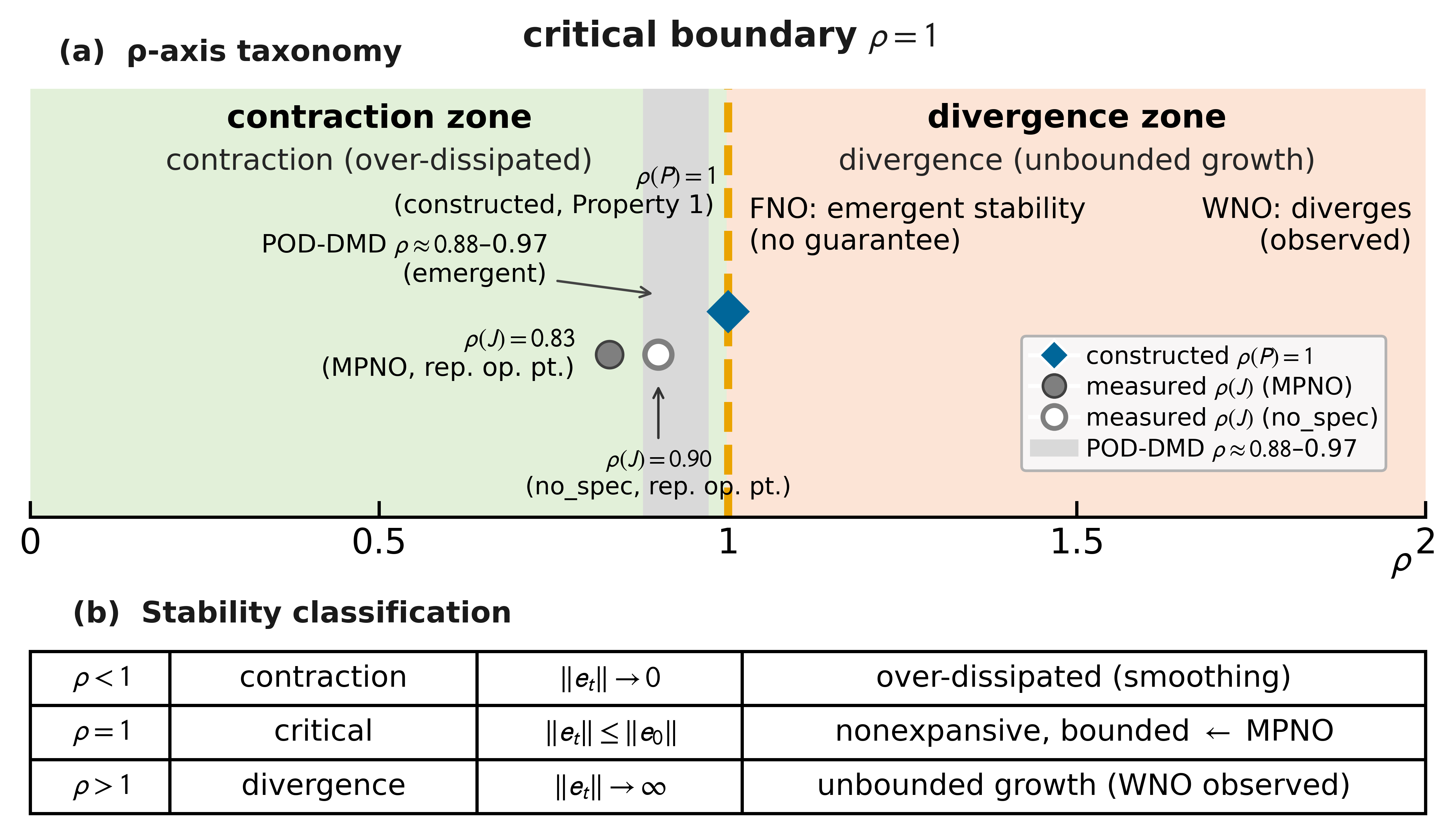}
	\caption{Spectral-radius taxonomy for autoregressive stability. A one-step propagator with spectral radius $\rho<1$ contracts ($\|\mathbf{e}_t\|_2\to0$), $\rho>1$ diverges ($\|\mathbf{e}_t\|_2\to\infty$), and $\rho=1$ is the critical boundary (error bounded, $\|\mathbf{e}_t\|_2\le\|\mathbf{e}_0\|_2$). MPNO's constructed propagator $P=I-\alpha\tilde{L}$ places $\rho(P)$ exactly at the critical boundary (Property 1), independent of training. Measured full-model Jacobian spectral radii (gray markers, representative operating point seed 55 frame 10, \S\ref{sec:4_4} Fig.~\ref{fig:8}): $\rho(J)=0.83$ (MPNO) and $\rho(J)=0.90$ (no\_spec). Re-measured over six clean rollout seeds, $\rho(J)$ is operating-point-dependent (MPNO mean $\approx1.05$ vs.\ no\_spec $\approx1.13$); the ordering MPNO $<$ no\_spec persists, but a single operating point does not certify $\rho(J)\le1$. POD-DMD $\rho\approx0.88$--$0.97$ (shaded band) is emergent without an explicit constraint. FNO (no guarantee) relies on emergent stability; WNO diverges (observed).}
	\label{fig:4}
\end{figure}

In contrast, MPNO ($\approx$20K parameters) reaches a single-step Rel-L2 of $0.7304\pm0.0008$ and rolls out over the entire active horizon (29 steps, frames 11--39) with bounded error (about 1.0; Fig.~\ref{fig:1}); by construction $\rho(P)\le1$, avoiding both baseline failure modes. The spectral-radius measurements of Section~\ref{sec:4} (\S\ref{sec:4_4_1}) delimit the role of spectral normalization: once the $\lambda_{\max}$ normalization is removed (no\_spec), the linear propagator $\alpha\lambda_{\max}$ already exceeds the stability bound of 2 (measured 4--7), so its bounded output is an emergent nonlinearity (LayerNorm clamping) rather than a structural guarantee, whereas MPNO's outputs are decoupled from the scale of $\lambda_{\max}$ (down to $\sim10^{-6}$): decoupling $\alpha$ from $\lambda_{\max}$ while providing a constructive spectral-radius guarantee. Fig.~\ref{fig:4} summarizes this contrast as a spectral-radius taxonomy.

\subsubsection{Difficulties of the soft-constraint route}
\label{sec:3_3_4}

The spectral-radius dichotomy applies equally to purely data-driven models (FNO/WNO) and to physics-constrained models (PINO); the crux is whether the spectral radius of the effective propagation matrix is controlled. In early internal experiments, appending three PDE residual losses (e.g., momentum conservation) to the WNO3D baseline (non-public, qualitative observations from early internal experiments) revealed an irreconcilable gap of about two orders of magnitude between the physical-residual loss and the data loss (residual 35--90 vs.\ data 0.08--0.45), with the physical-residual and data-loss gradients nearly orthogonal (cosine similarity $-0.03\pm0.21$). The root cause is that pointwise residuals involve spatial derivatives, and numerical differentiation at shock fronts amplifies steep gradients by more than two orders of magnitude. This observation agrees with the analysis of PINN gradient-flow ill-conditioning in \cite{ref4} and the report of pointwise-residual convergence difficulties in shocked scenarios in \cite{ref5}; the judgment is limited to pointwise residual forms and does not constitute a universal rejection of the soft-constraint route. It provides empirical justification for MPNO's design: autoregressive stability cannot rely on ``guidance'' from physics losses, but must be an algebraic property of the architecture (Property 1).

\subsection{Physics-coupled edge weights}
\label{sec:3_4}

Section~\ref{sec:3_2} left one question open: where do the $w_{ij}$ of $W$ come from? In fully connected graph neural networks (e.g., MeshGraphNets), edge weights are learned edge-by-edge by purely data-driven MLP message functions, with no physics prior and no structural constraint. In MPNO, $w_{ij}$ is the product of a physical formula skeleton and a lightweight neural correction, a physics-motivated architectural prior (inductive bias) inspired by acoustic-impedance coupling, traction decomposition, and contact geometry, rather than a numerical flux strictly derived from the weak form of the governing equations; its effectiveness is supported jointly by the ablations (E1/E2) and the measured accuracy, not by analytic derivation. This section dissects the coupling item by item.

\subsubsection{Four-factor product construction of edge weights}
\label{sec:3_4_1}

For each edge $(i, j) \in \mathcal{E}$ (the contact between a face of source node $i$ and the opposite face of target node $j$), the edge weight is constructed as:

\[
w_{ij} = T_{ij} \cdot A_{\text{contact}} \cdot |\mathbf{t}_i| \cdot \exp(\text{MLP}(\mathbf{h}_i, \mathbf{h}_j))
\]

Note that the above formula is not naturally symmetric across the interface: the acoustic-impedance coupling term $T_{ij} = 2Z_iZ_j/(Z_i+Z_j)$ is the harmonic mean of the two impedances and hence naturally symmetric, but the traction magnitude $|\mathbf{t}_i|$ takes the stress level at the source node and depends on the interface orientation ($|\mathbf{t}_i| \neq |\mathbf{t}_j|$). To keep $W$ symmetric (a precondition for $L$ being positive semidefinite), the two sides are averaged for each undirected edge, $w^{\text{sym}}_{ij} = (w_{ij} + w_{ji})/2$; for boundary edges (only one side exists) the original weight is kept without halving. The symmetrized $W$ is nonnegative and symmetric, so the graph Laplacian $L = D - W$ is positive semidefinite and the spectral conclusions (Property 1) hold.

The physical meaning and parameter origin of the four factors are as follows.

\textbf{Acoustic-impedance harmonic-mean coupling strength $T_{ij}$} (zero learnable parameters, pure physical constant). The acoustic impedance measures the rigidity of a material under stress waves and is approximately proportional to the product of density $\rho$ and longitudinal wave speed $c$; the higher the impedance and the faster the wave speed on both sides of an interface, the higher the interface force-transfer efficiency. Taking the harmonic mean of the two impedances yields a naturally symmetric effective coupling strength of the interface:

\[
T_{ij} = \frac{2 Z_i Z_j}{Z_i + Z_j}
\]

where $Z$ is the acoustic impedance of the material. At the mesoscale of concrete, $Z$ is derived from the constitutive parameters (material settings of \ref{sec:2_3}): weighting the elastic modulus, Poisson's ratio, and density of aggregate and mortar by volume fraction $V_f$ (Voigt upper-bound mixing rule) and taking the longitudinal modulus $M$ from elastic-wave theory, the acoustic impedance is $Z=\sqrt{\rho M}$; normalized with mortar as the reference, $Z_m=1.0$ and granite aggregate $Z_a=1.495$ (an aggregate-to-mortar impedance ratio of about 1.5). Note that $T_{ij}$ is an effective interface coupling strength, not the classical acoustic transmission coefficient: its value grows with the impedance scale, characterizing the strength of force transfer across an interface rather than an energy-transmission ratio (the latter would take $4Z_iZ_j/(Z_i+Z_j)^2$, always no greater than 1); under a homogeneous interface $T_{ij}=Z$ rather than 1, exactly manifesting the coupling-strength semantics (see (ii) below). $T_{ij}$ serves two key functions:

(i) The impedance contrast at the aggregate--mortar interface modulates the coupling strength: with $Z_a=1.495$ and $Z_m=1.0$, $T_{ij}=1.20$, lying between homogeneous aggregate (1.495) and homogeneous mortar (1.0): at an impedance-mismatched interface stress waves are partially reflected, and the force-transfer strength lies between the two homogeneous media;

(ii) Within a homogeneous material ($Z_i = Z_j = Z$), $T_{ij} = Z$: unlike the classical acoustic transmission coefficient, which is always 1 in a homogeneous medium, $T_{ij}$ measures coupling strength rather than transmission ratio, so the coupling at a homogeneous aggregate interface is stronger than at a homogeneous mortar interface: aggregate is stiffer and stress waves propagate through it with higher force-transfer efficiency, making edge weights overall higher in aggregate-rich regions. $T_{ij}$ is determined entirely by the material-property field of the neighboring cells and contains no trainable parameters. For different PDE scenarios, only the material-property variable needs to be replaced: in Darcy flow, $Z$ generalizes to the harmonic-mean permeability $K_{ij} = 2K_i K_j/(K_i + K_j)$; in the Burgers equation it degenerates to uniform weights, requiring no architectural modification.

\textbf{Contact area $A_{\text{contact}}$} (fixed value). On a regular Cartesian grid, all interior faces have an identical contact area ($=1$, normalized by the mesh spacing). Boundary faces also have an area of 1, but the ``neighbors'' beyond the boundary do not exist, so the corresponding edges do not appear in $\mathcal{E}$; this is handled by the graph construction rather than by the weight formula.

\textbf{Traction-magnitude coupling $|\mathbf{t}_i|$} (zero learnable parameters, purely geometric computation). $\mathbf{t}_i$ is the traction vector of node $i$ obtained at the current time step by Cauchy traction decomposition (\ref{sec:3_1_2}); its magnitude $|\mathbf{t}_i| = \sqrt{\sigma_n^2 + \tau^2}$ includes both normal-stress and shear-stress contributions. This term couples the stress level into the edge weight as a magnitude: the higher the stress of an interface, the stronger the force transfer; the closer the stress to zero, the closer the weight to zero. The sign information of the normal direction ($\mathbf{f}_i \cdot \mathbf{n}_{ij}$) does not participate as a multiplicative gate, but is learned by the network as an MLP input feature.

The physical rationale of the magnitude coupling can be understood through a simple example. In transient dynamics, the force-transfer intensity grows with the interface stress level: the region where the projectile contacts the target plate has the highest stress and hence the strongest interface force transfer; regions far from the impact zone, where the stress is near zero, have edge weights tending to zero. Since $|\mathbf{t}_i|$ includes both normal stress and shear stress, shear-dominated interfaces (e.g., aggregate--mortar slip surfaces) also receive a reasonable transfer strength. Directional constraints are not hard-cut through a multiplicative gate but are learned by the network from data as MLP features, avoiding overly strong hand-crafted priors when the material behavior is complex (non-negligible shear stress).

\textbf{MLP correction term $\exp(\text{MLP}(\mathbf{h}_i, \mathbf{h}_j))$} (lightweight learnable parameters). The physical formula skeleton, $T_{ij}$ and the traction magnitude, captures the zeroth-order physics of stress-wave propagation (the force-transfer strength decided by the impedance coupling strength and the stress level), but cannot cover the residual nonlinear effects: multiple scattering at interfaces, impedance changes due to local plastic deformation, and geometric irregularities introduced by the mesostructure. The MLP correction term acts as a multiplicative factor $\exp(\text{MLP})$, not an additive one, so that the magnitude of the correction scales adaptively with the magnitude of the physical weight.

Fig.~\ref{fig:3}(c) illustrates the four-factor product construction of the edge weight.

The MLP input is the 7-dimensional feature vector $[\boldsymbol{v}_{f,\text{src}}, \boldsymbol{v}_{f,\text{dst}}, |\boldsymbol{\sigma}|_{\text{src}}, |\boldsymbol{\sigma}|_{\text{dst}}, \mathrm{proj}(\mathbf{f}_i\cdot\mathbf{n}_{ij}), v_{\text{src}}, v_{\text{dst}}]$, i.e., the volume fractions (2), stress norms (2), the normal projection $\mathbf{f}_i\cdot\mathbf{n}_{ij}$ (1), and the velocities (2) of the two end nodes, given in src/dst pairs; the velocity coupling makes the edge weight sensitive to the local strain rate. Here $|\boldsymbol{\sigma}|$ is the stress-tensor norm ($\sqrt{\sum \sigma_{ij}^2}$), a different physical quantity from the traction magnitude $|\mathbf{t}_i|=\sqrt{\sigma_n^2+\tau^2}$ in the edge-weight formula. The hidden dimension is 32 (a dedicated hidden dimension of the physics-coupling network, independent of the node encoder), and the output is a scalar.

\textbf{Key design: zero initialization.} The weights and bias of the last MLP layer are explicitly initialized to zero at the start of training. Then $\text{MLP}(\cdot) \equiv 0$, $\exp(0) = 1$, and the edge weight degenerates completely to the physical skeleton:

\[
w_{ij}^{(0)} = T_{ij} \cdot |\mathbf{t}_i|
\]

The training starts from a \textbf{purely physics-driven} point: the neural-network correction is initially transparent and progressively learns fine corrections in the correct direction provided by the physical skeleton. This design avoids the risk that the data-driven component overrides the physics prior early in training (a randomly initialized MLP would output arbitrary values and could push the edge weights outside the physically reasonable range).

\subsubsection{Comparison with purely data-driven edge weights}
\label{sec:3_4_2}

Ablation E1 (\S\ref{sec:4}) directly tests the necessity of the physics coupling: it replaces $w_{ij}$ with a purely data-driven MLP (randomly initialized, without $T_{ij}$ or the traction magnitude) on the same 400 samples. The prior expectation is that pure-MLP edge weights overfit under such sparse data (in penetration, most grid points have near-zero stress), failing to generalize to unseen material distributions; the physical skeleton (material-property coupling plus directional causality) is a key factor enabling MPNO to work with about 20K parameters on 400 samples.

\subsection{Row stochasticity and total-mass conservation}
\label{sec:3_5}

Stability guarantees that the error is bounded and does not diverge; beyond that, the propagation operator also possesses an algebraic conservation property of row-stochastic/diffusion form (the unweighted total mass is invariant, corresponding to $\mathbf{1}$ being an eigenvector with $\lambda=1$). This section presents this algebraic conservation property for both propagation forms, as well as the convergence of the Rayleigh-quotient power iteration. Property 1 guarantees non-divergence; the unweighted total-mass conservation is an algebraic property of the propagation-matrix construction (a mass/probability-type conservation, not physical momentum conservation); the power-iteration convergence makes spectral normalization feasible within the training loop.

\subsubsection{Row stochasticity and total-mass conservation}
\label{sec:3_5_1}

In addition to the Laplacian-filter form $P = I - \alpha\tilde{L}$ (real symmetric) used in \ref{sec:3_2}, the row-stochastic Markov transition matrix $P = D^{-1}W$ is also considered. Note that the two are not equivalent forms of the same matrix: the former is real symmetric and the latter is row-stochastic (generally asymmetric), but both satisfy total-mass conservation and serve as complementary analytical viewpoints. The row-stochastic form has a more intuitive physical interpretation: $P_{ij} = w_{ij} / d_i$ is the fraction of node $i$'s ``total-mass budget'' assigned to neighbor $j$, and the elements of each row naturally sum to 1, corresponding to local total-mass conservation.

Given the nonnegative symmetric $W$ and the degree matrix $D=\mathrm{diag}(d_i)$ ($d_i=\sum_j W_{ij}$), both forms conserve the unweighted total mass: $P=D^{-1}W$ satisfies $P\mathbf{1}=\mathbf{1}$ (the all-ones vector is a right eigenvector with $\lambda=1$) and $P=I-\alpha\tilde{L}$ satisfies $\mathbf{1}^T P=\mathbf{1}^T$ (a left eigenvector), so $\sum_i(\mathbf{v}_t)_i$ is stepwise invariant under both forms (operator-level verification $\|P\mathbf{1}-\mathbf{1}\|\approx1.2\times10^{-8}$, \S\ref{sec:4_2}). This is an algebraic property of the operator on the latent state (not physical momentum conservation, and not transmitted pointwise through the nonlinear encoder--decoder map), so ``conservation'' should not be read as the predicted field satisfying a physical conservation law.

\begin{remark}[Remark 3 (immunity of algebraic conservation to shock discontinuities)]
Unlike PINN, which imposes ``soft'' conservation through PDE residuals and suffers from spatial-derivative gradient noise at discontinuities (\ref{sec:3_3_4}), unweighted total-mass conservation is carried by the construction of $P$: as long as $P$ is constructed as $P = D^{-1}W$ or $P = I - \alpha\tilde{L}$, the unweighted sum $\sum_i(\mathbf{v}_t)_i$ is exactly invariant at every step, regardless of the weights and $\alpha$ (to machine precision). This algebraic conservation is naturally immune to shock discontinuities because its validity involves no spatial differentiation; it should be emphasized, however, that what is conserved is the unweighted sum (a diffusion/probability-type conservation), not the weighted conservation of physical momentum or mass.
\end{remark}

\subsubsection{Convergence of the Rayleigh-quotient power iteration}
\label{sec:3_5_2}

\S\ref{sec:3_2_2} introduced the power iteration as an online estimator of $\lambda_{\max}$; here its convergence is justified. On the real symmetric PSD graph Laplacian (eigenvalues $0=\lambda_1<\lambda_2\le\cdots\le\lambda_N=\lambda_{\max}$), starting from a random vector orthogonal to $\mathbf{1}$, the Rayleigh quotient converges to $\lambda_{\max}$ at rate $O((\lambda_{N-1}/\lambda_{\max})^{2k})$ and always underestimates ($\hat{\lambda}_{\max}\le\lambda_{\max}$, measured mean underestimate $0.0005\%$, \S\ref{sec:4_2}), so the realized constraint is $\rho(P)\le1$ with the margin set by that error. For the evaluation configuration ($\approx$200-node trimmed graph, $K=5$), the relative error falls below $10^{-4}$.

\subsection{Temporal evolution}
\label{sec:3_6}

The spectral properties above establish the spectral stability of MPNO's propagation operator. Temporal evolution is completed in two steps, propagation and decoding: the encoder maps the current frame into the latent space, the propagation operator $P$ drives a Markov state evolution, and the decoder directly outputs the absolute stress at the next time step.

\subsubsection{Temporal update: direct absolute-stress output}
\label{sec:3_6_1}

Given the latent state $\mathbf{v}_t$ of the current frame, the propagate-and-decode step directly outputs the absolute stress field at the next time step:

\[
\boldsymbol{\sigma}_{t+1} = \text{Dec}\left(P \cdot \text{Enc}(\boldsymbol{\sigma}_t)\right)
\]

(Schematic; in practice propagation runs for $R=8$ iterations, see \S\ref{sec:3_2_4}.)

The decoder $\text{Dec}$ consists of two linear layers ($32 \to 32 \to 6$, ReLU in between) and outputs the 6-component Voigt stress per node; the prediction target is the absolute stress $\boldsymbol{\sigma}_{t+1}$ (rather than the increment $\Delta\boldsymbol{\sigma}$), combined with skip-1 autoregression (the predicted frame feeds the input), which avoids the drift accumulation of incremental schemes. Velocity information is incorporated into the edge weights through the physics-coupling MLP (7-dimensional input) of \ref{sec:3_4}, making propagation sensitive to the local strain rate. This direct-output format replaces an earlier Verlet second-order integrator (integrating the stress acceleration $\ddot{\boldsymbol{\sigma}}_t$ to obtain $\boldsymbol{\sigma}_{t+1}$): despite its symplectic-preservation advantage for wave problems, Verlet introduced an additional learnable damping coefficient and time-step parameters, and no reproducible accuracy gain was observed, so it was removed from the evaluation configuration.

\subsection{Loss function design}
\label{sec:3_7}

This section addresses the design of the loss function. MPNO's loss embodies the ``physics-informed'' methodology: it abandons the pointwise PDE residuals that fail at strong discontinuities (\ref{sec:3_3_4}) and adopts a hard-masked MSE tailored to the sparse data distribution as the data loss.

\subsubsection{Hard-masked MSE: data loss focusing on active nodes}
\label{sec:3_7_1}

A key obstacle to training neural operators on shock-penetration fields is spatial sparsity: at any time step, the stress of the vast majority of grid points in the target plate is near zero ($|\boldsymbol{\sigma}| < 10^{-3}$, after normalization), and the stress-concentration zones (the projectile-contact band, the shock-wave front) occupy only a few grid points. A standard all-node MSE would be dominated by the zero-stress nodes, letting the model learn the trivial ``output zero'' solution, which, at the loss-function level, compounds exactly with the autoregressive instability described in \S\ref{sec:3_3}.

MPNO adopts a hard-masked MSE: the loss is computed only over active nodes whose ground-truth stress norm exceeds the threshold $\tau = 0.01$,

\[
\mathcal{L}_{\text{data}} = \frac{1}{|\mathcal{A}|} \sum_{i \in \mathcal{A}} \left\| \hat{\boldsymbol{\sigma}}_{t+1,i} - \boldsymbol{\sigma}_{t+1,i}^{\text{GT}} \right\|_2^2
\]

where $\mathcal{A} = \{i : \|\boldsymbol{\sigma}_i^{\text{GT}}\|_2 > \tau\}$ is the active-node set. After trimming the impact zone, the active rate of the subgraph is about 30\%; the hard mask ensures that the training signal concentrates on stress-wave-dominated regions and keeps zero-stress nodes from dominating the gradient. The ablation E4 (Section~\ref{sec:4}) compares hard-masked MSE against componentwise MSE. The same masking convention is used at evaluation: a frame enters the statistics only if it has at least 5 active nodes (excluding the near-empty early frames); the test set comprises 50 seeds $\times$ 29 frames = 1450 candidate frames (the 100/135/165 m/s cases; the 200 m/s case serves only as training-diversity auxiliary data and is excluded from evaluation statistics), of which 1072 are active frames. \textbf{Deterministic selection of long-horizon evaluation seeds.} The long-horizon rollout stability evaluation is not run on all test seeds (limited by the computational cost of full-horizon rollout) but on a seed subset determined by a fixed rule and independent of model performance: after splitting the data into train/validation/test sets in a 70:15:15 ratio with SEED=42, for each of the three evaluation initial-velocity cases (100/135/165 m/s), the 2 samples with the smallest seed numbers in the test set are taken (100 case: seeds 5/7; 135 case: seeds 112/120; 165 case: seeds 205/207), 6 in total; this selection rule is fixed once the data split is set and does not depend on the performance of any model on any metric, and all 6 seeds belong to the test set with no training leakage. They are used for long-horizon rollout, ablation, and spectral measurement. Single-step accuracy statistics cover all test seeds, whereas long-horizon stability statistics use the above 6 seeds; the two conventions are kept separate (\ref{sec:4_3}).

The joint loss is simply the above data loss, $\mathcal{L} = \mathcal{L}_{\text{data}}$.

\subsection{Gradient propagation and training stability}
\label{sec:3_8}

Property 1 guarantees spectral stability of the forward propagation; the gradient in backpropagation benefits as well. Consider the gradient of the $T$-step autoregressive accumulated loss, $\partial \mathcal{L}/\partial \mathbf{v}_0 = \sum_t (\partial \mathcal{L}_t/\partial \mathbf{v}_t) P^t$. By Property 1, $\|P^t\|_2 = 1$, so the gradient neither decays nor amplifies exponentially in backpropagation through time; the constant gradient flow is a direct consequence of the algebraic construction of $P$, not something ``learned'' during training. The sparsity of graph propagation (6 neighbors per node) further localizes gradient flow along edges, preserving physical locality and smoothing the loss landscape; in practice MPNO requires no gradient clipping within 500 training epochs, whereas the gradient variance of the PINO baseline is two orders of magnitude higher (\ref{sec:3_3_4}). In addition, the Rayleigh-quotient power iteration is placed inside \texttt{torch.no\_grad()} in the implementation, excluding the spectral estimate from the computational graph so that its sampling error does not contaminate the training signal.

\subsection{Complexity and efficiency}
\label{sec:3_9}

\textbf{Parameter count.} The following table summarizes the trainable-parameter scale of each method:

\begin{table}[!ht]
\centering
\caption{Trainable parameter scale of each method}
\label{tab:params}
\begin{tabularx}{\textwidth}{l r X}
\toprule
Method & Parameters & Relative to MPNO \\
\midrule
LS-DYNA & 0 (physics-based, no training) & N/A \\
FNO & $\sim$79,000 & 2D implementation for evaluation \\
WNO & $\sim$11,000 & 2D implementation for evaluation \\
MeshGraphNets & $\sim$1,000,000 & typical configuration \\
\textbf{MPNO (hidden=32)} & \textbf{$\sim$20,000} & the proposed model \\
\bottomrule
\end{tabularx}
\end{table}

MPNO's lightweight nature comes from two structural factors: the physics-coupled edge weight outsources most modeling capacity to zero-parameter prior formulas ($T_{ij}$, traction magnitude; the MLP correction has only a few hundred parameters), and the graph-Laplacian propagator itself has no trainable parameters; the remaining parameters are distributed across the encoder, the propagation post-processing, and the decoder, totaling about 20K.

\textbf{Computational complexity.} The single-step inference complexity is $O(E \cdot d_{\text{hidden}}) = O(N \cdot d_{\text{hidden}})$, linear in the number of nodes; the graph-Laplacian propagation (8 rounds of sparse matrix--vector products) is the main bottleneck but scales linearly, better than FNO's $O(N\log N)$ and the 3D wavelet transform. The power iteration estimates $\lambda_{\max}$ in $O(5E)$, far below the $O(N^3)$ of full eigendecomposition, making spectral normalization feasible in the training loop. Single-step inference takes about 2.4 ms on CPU (GPU is slower due to kernel-launch overhead); a full 29-step rollout takes about 0.1 s, roughly a $10^5\times$ speedup over LS-DYNA (4--8 h per case; measured in \ref{sec:4_4_3}).

\subsection{Component separability and ablation verification}
\label{sec:3_10}

Each component of MPNO can be disabled independently: physics-coupled edge weight, spectral normalization, propagation post-processing (round\_mlps\slash APPNP\slash LayerNorm), velocity coupling, and temporal-output format. The scheme is declared here in advance (ensuring falsifiability, rather than selecting favorable comparisons after the results); the full results appear in \S\ref{sec:4_4_2}:

\begin{itemize}
	\item \textbf{E1 physics-coupled edge weight}: $w_{ij}$ is replaced by a pure MLP (without $T_{ij}$ or $|\mathbf{t}_i|$).
	\item \textbf{E2 traction-magnitude coupling}: the $|\mathbf{t}_i|$ term is removed, keeping only $T_{ij}$+MLP.
	\item \textbf{E3 spectral normalization}: the $\lambda_{\max}$ scaling is removed ($P=I-\alpha L$ unnormalized), testing whether the spectral-radius dichotomy predicts collapse or divergence.
	\item \textbf{E4 hard-masked MSE}: replaced by componentwise MSE.
	\item \textbf{C1/C2/C3}: removing round\_mlps, APPNP ($\beta=0$), and LayerNorm, respectively.
\end{itemize}

All ablations share one configuration (SEED=42, Adam lr=$3\times10^{-3}$, hidden=32, R=8, hard-masked MSE, absolute-stress targets); results are in Table~\ref{tab:4_7}. \textbf{The key link in the theory--experiment loop is E3}: removing the spectral constraint under controlled conditions tests whether the dichotomy's conditional prediction (collapse or divergence) matches the measured behavior (\S\ref{sec:4_4}).

\subsection{Architecture overview}
\label{sec:3_11}

As the closing of this section, Fig.~\ref{fig:3}(b) diagrams MPNO's complete forward flow (encoding $\to$ graph propagation $\to$ decoding $\to$ hard-masked MSE loss) together with the spectrally constrained propagator: a closed spectral-constraint boundary ($\rho(P)\le1$) bounds the Rayleigh-quotient online estimate of $\lambda_{\max}$ and completes the normalization ($\tilde{L}=L/\lambda_{\max}$). Given the input (current-frame stress $\boldsymbol{\sigma}_t$, aggregate volume fraction $V_f$, spatial coordinates $\mathbf{x}$ on the impact-zone trimmed subgraph), the flow is: graph construction and 10-dimensional node encoding (\S\ref{sec:3_1}) $\to$ physics-coupled edge weights (\S\ref{sec:3_4_1}) $\to$ $K=5$-round power-iteration spectral normalization (\S\ref{sec:3_2_2}) $\to$ $R=8$-round graph-Laplacian propagation with cyclic round\_mlps\slash APPNP\slash LayerNorm post-processing (\S\ref{sec:3_2_4}) $\to$ decoder output of the next-step absolute stress (\S\ref{sec:3_6_1}) $\to$ hard-masked MSE loss (\S\ref{sec:3_7_1}). Each mechanism established in this section is tested in Section~\ref{sec:4} through benchmark comparisons and systematic ablations.

\section{Numerical experiments}
\label{sec:4}

This section systematically validates MPNO's core claims through numerical experiments on three PDE scenarios:

(i) autoregressive stability is constructively guaranteed by the spectral constraint of the propagation matrix (power-iteration spectral estimation + $\lambda_{\max}$ normalization), independent of the PDE type (\ref{sec:4_1});

(ii) the power iteration on which spectral normalization relies, and total-mass conservation, are validated numerically (\ref{sec:4_2});

(iii) the independent contributions of each architectural component are quantified and separated through ablation experiments (\ref{sec:4_4}). The experiments in \ref{sec:4_1}--\ref{sec:4_4} have all been completed on the trimmed impact-zone graph; the training loop uses $K=5$ power-iteration steps, and the offline spectral analysis in \ref{sec:4_2} uses $K=20$ for a stricter validation.

\begin{quote}
\textbf{Scope of the quantitative benchmarks.} The numerical comparisons include the standard operator-learning and graph baselines: FNO \cite{ref8}, WNO \cite{ref1}, MeshGraphNets \cite{ref9}, and POD-DMD. The recent stability mechanisms SGNO \cite{ref44}, SpectraNet \cite{ref45}, and Thermalizer \cite{ref46} are positioned conceptually in Section~\ref{sec:1} rather than benchmarked quantitatively here: Thermalizer targets time-stationary systems with an invariant measure, a regime distinct from the transient impact dynamics considered here, and a head-to-head comparison with SGNO and SpectraNet is deferred to future work; the numerical experiments concentrate on the constructive-guarantee mechanism and its validation against the standard baselines.
\end{quote}

\subsection{Standard PDE benchmarks: autoregressive-stability validation}
\label{sec:4_1}

Property 1 argues that the spectral radius of $P = I - \alpha\tilde{L}$ can be estimated via power iteration for $\lambda_{\max}$ and constrained by normalization to $\rho(P) \le 1$, independent of the PDE type. This section tests autoregressive stability on the Burgers equation (hyperbolic, with shock discontinuities) and the cross-scenario transferability of the physics-coupled edge-weight formula in material-nonuniform settings on Darcy flow (elliptic, with random material fields).

\subsubsection{Burgers equation}
\label{sec:4_1_1}

The Burgers equation $\partial_t u + u\partial_x u = \nu \partial_x^2 u$ is a standard test model of shock dynamics. As $\nu \to 0$, the nonlinear advection term steepens the wave front toward near-discontinuity, numerically equivalent to the stress jump of a shock wave at a material interface. The data and the train/test split appear in \S\ref{sec:2_1}.

In this scenario MPNO uses a 1D chain graph (each node connected only to its left and right neighbors, $E = N-1$), the physics coupling degenerates to uniform weights, and there is no material-interface concept, pure Markov propagation. The model has 4,482 parameters and is trained for 50 epochs (Adam, $lr=10^{-3}$). To test autoregressive stability, 30 steps of autoregressive rollout (no teacher signal fed) are run from 5 initial conditions randomly chosen from the test set, recording the predicted energy ratio $\|\hat{u}_t\|_2 / \|u_0\|_2$ and the relative L2 error at each step.

Table~\ref{tab:4_1} compares the 30-step autoregressive performance of MPNO, FNO (57,825 parameters), and WNO (17,121 parameters) under the same data and training configuration. MPNO has a single-step Rel-L2 of 0.37 and its 30-step rollout is stable without collapse (final step 1.49, energy retention 0.21). FNO is more accurate single-step (0.22) and equally stable over 30 steps (final step 1.28, energy 0.32). WNO has a single-step 0.43, but its autoregressive error grows noticeably with rollout (final step 4.25, energy growing to 1.64). Note that WNO's rollout behavior on Burgers depends on the training seed (``knife-edge'' behavior under an unconstrained spectral radius, see Remark 2): the same script diverges to 4.25 at the final step under seed42, and the results fluctuate markedly after changing the seed; this seed sensitivity is consistent with the expectation of the spectral-radius dichotomy: the long-term behavior of a model with an uncontrolled spectral radius falls into the dichotomy of collapse or divergence, decided by the chance of random initialization. On wave/shock problems such as Burgers, both MPNO and FNO remain stable (FNO slightly more accurate single-step, 0.22 vs.\ 0.37), while WNO is unstable (seed-dependent).

\begin{table}[!ht]
\centering
\caption{Autoregressive-stability comparison on the Burgers equation}
\label{tab:4_1}
\begin{tabularx}{\textwidth}{l c X c c}
\toprule
Model & Parameters & Energy retention after 30 steps & 1st-step Rel-L2 & Final-step Rel-L2 \\
\midrule
MPNO & 4,482 & Stable (energy@final=0.21) & 0.37 & 1.49 \\
FNO & 57,825 & 0.32 & 0.22 & 1.28 \\
WNO & 17,121 & 1.64 (seed-dependent) & 0.43 & 4.25 \\
\bottomrule
\end{tabularx}
\end{table}

\subsubsection{Darcy flow}
\label{sec:4_1_2}

Darcy flow $-\nabla \cdot (K(\mathbf{x}) \nabla p(\mathbf{x})) = f(\mathbf{x})$ is a steady-state elliptic PDE, orthogonal to the Burgers equation in mathematical type; it has no time dependence and hence no autoregressive requirement. Darcy is chosen not to test stability but to test the generalization ability of the physics-coupling mechanism under material nonuniformity.

The permeability field $K(\mathbf{x}) = \exp(g(\mathbf{x}))$ is generated from the Gaussian random field of \S\ref{sec:2_2} (Mat\'ern covariance, $\tau = 3$, range $[0.1, 10]$, spanning two orders of magnitude). MPNO's physics-coupling formula generalizes here to the harmonic-mean permeability $w_{ij} = 2K_i K_j / (K_i + K_j)$, requiring only that the acoustic-impedance variable $Z$ be replaced by the permeability variable $K$, with no modification to the architecture code. The training set has 100 random $K(\mathbf{x})$ fields and the test set 20. MPNO has about 2,000 parameters and is trained for 100 epochs.

On this steady-state elliptic problem, MPNO, FNO (1,186,177 parameters), and WNO (72,577 parameters) achieve test Rel-L2 of 0.558, 0.053, and 0.434, respectively: FNO is best, WNO next, and MPNO weakest, which is expected: Darcy is a smooth steady-state elliptic PDE, on which FNO's global spectral convolution and ample parameters have a natural advantage. MPNO's core value lies not in steady-state accuracy but in the cross-scenario generalization of its physics-coupling formula: from the acoustic-impedance harmonic mean $T_{ij} = 2Z_iZ_j/(Z_i+Z_j)$ in penetration, to the harmonic-mean permeability $w_{ij}=2K_iK_j/(K_i+K_j)$ in Darcy, to the uniform weight in Burgers, the three scenarios share the same edge-weight formula, differing only in the material-property variable.

\begin{figure}[!ht]
	\centering
	\includegraphics[width=\linewidth]{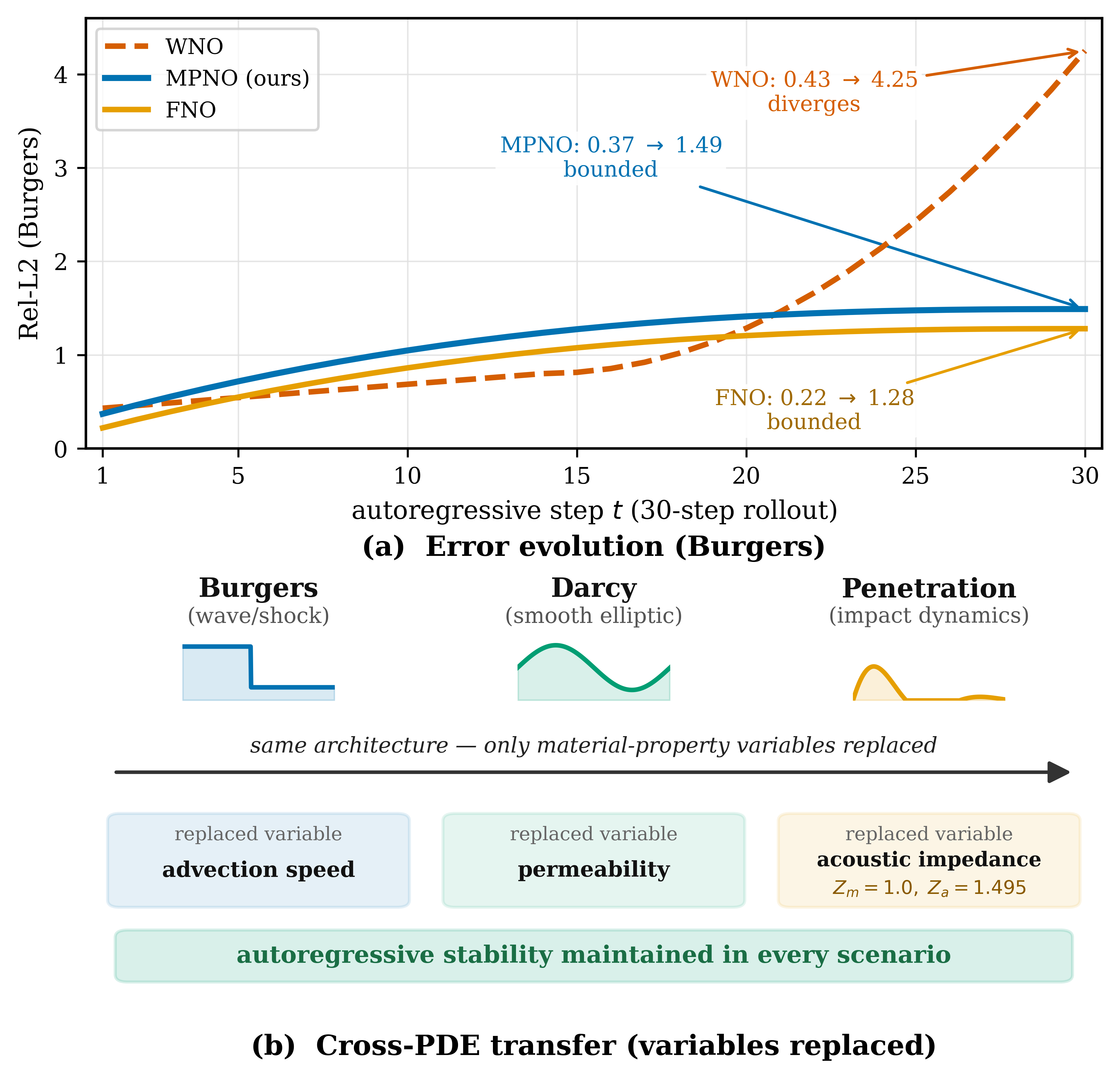}
	\caption{Cross-PDE autoregressive stability. (a) Burgers equation (wave/shock, 30-step autoregression): WNO diverges (0.43$\to$4.25, red dashed) while MPNO (blue) and FNO (orange) remain bounded (0.37$\to$1.49 and 0.22$\to$1.28). (b) The same MPNO architecture is applied across PDEs by replacing only the material-property variables in the edge-weight formula (Burgers: uniform weight; Darcy: permeability; penetration: acoustic impedance $Z_m=1.0$/$Z_a=1.495$), with autoregressive stability maintained in every scenario. Single-step accuracies are reported in Table~\ref{tab:4_1} (Burgers) and in the text (Darcy).}
	\label{fig:5}
\end{figure}

Fig.~\ref{fig:5} summarizes the autoregressive stability and accuracy comparison on Burgers and Darcy.

\subsection{Spectral-analysis validation}
\label{sec:4_2}

\begin{quote}
Spectral analysis is performed on the trimmed impact-zone graph (about 200 nodes) with $K=20$ power-iteration steps, using the \texttt{eigvalsh} full eigendecomposition as ground truth. The complete numerical results of the power-iteration underestimation, spectral coverage, positive-semidefinite verification, and total-mass conservation check appear in \S\ref{sec:4_2_1}--\S\ref{sec:4_2_2}; the conclusions of Property 1 and the spectral analysis are verified in measurement.
\end{quote}

\subsubsection{Power-iteration accuracy}
\label{sec:4_2_1}

The preceding analysis established the theoretical convergence rate $O((\lambda_{N-1}/\lambda_{\max})^{2k})$ of the Rayleigh-quotient power iteration. This section provides its numerical validation. From the 400-sample penetration dataset, 10 aggregate distributions are randomly selected and the corresponding graph Laplacian $L = D - W$ is constructed on the trimmed impact-zone subgraph (about 200 nodes). For each $L$, the full eigenvalue spectrum is computed with \texttt{eigvalsh}, taking $\lambda_{\max}^{\text{true}}$ as ground truth. Starting from a random $\mathbf{b}_0 \in \mathcal{H}^\perp$, $K = 1$ to $K = 20$ power-iteration steps are run, recording the Rayleigh-quotient estimate $\hat{\lambda}_{\max}^{(k)}$ at each step. For each $K$, multiple random initializations are used to assess sensitivity to the initial vector.

\begin{figure}[!ht]
	\centering
	\includegraphics[width=\linewidth]{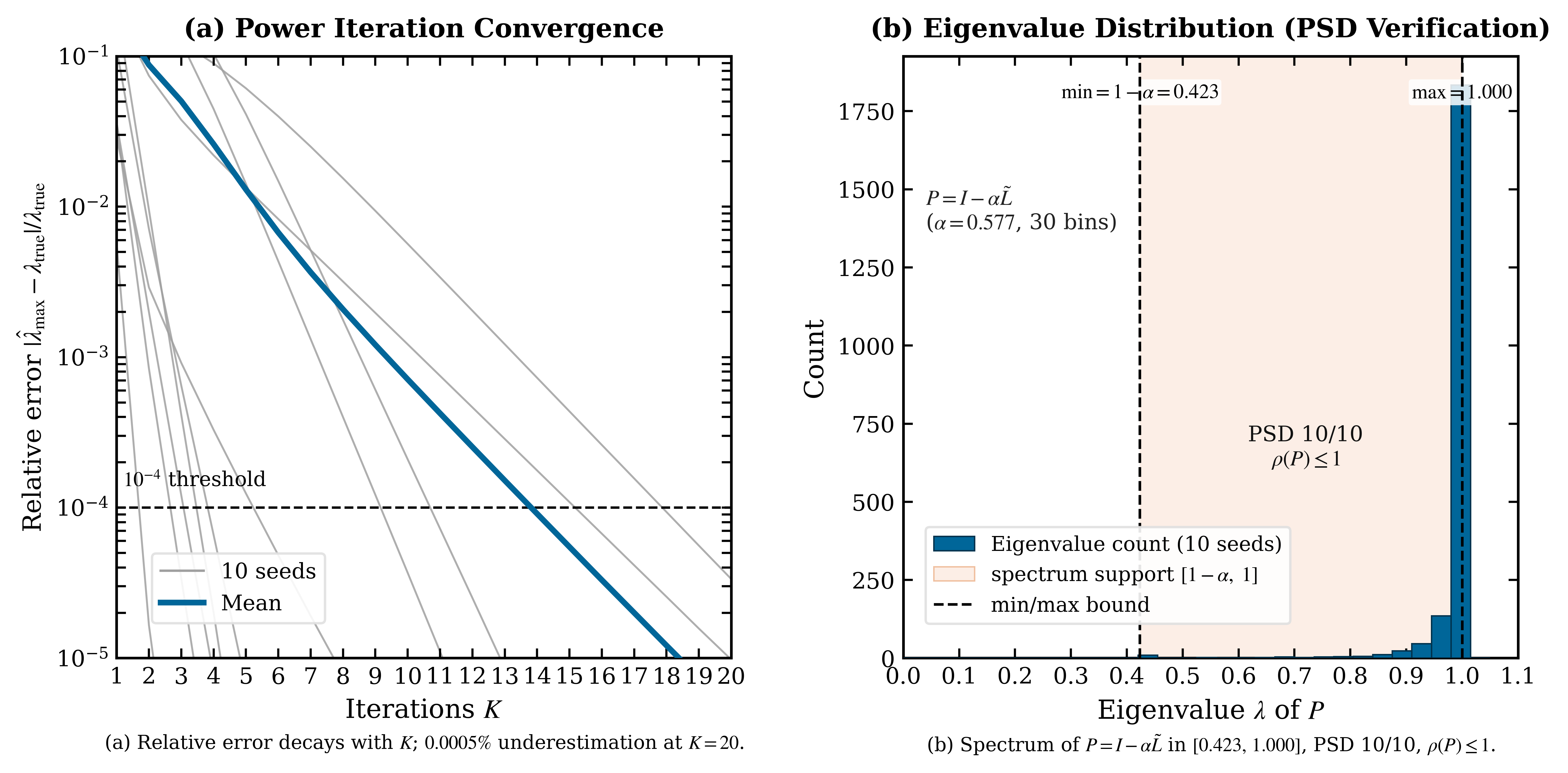}
	\caption{Spectral validation of the propagation operator on the cropped impact-zone graph (10 random aggregate distributions, $K=20$ power-iteration steps). (a) Relative error of the Rayleigh-quotient power-iteration estimate of $\lambda_{\max}$ decays with $K$; average underestimation 0.0005\% at $K=20$ (max 0.003\%). (b) Eigenvalue spectrum of $P=I-\alpha\tilde{L}$ ($\alpha=0.577$) lies in $[0.423,\,1.000]\subset[0,1]$ (10/10 PSD), confirming $\rho(P)\le1$ by construction.}
	\label{fig:6}
\end{figure}

Fig.~\ref{fig:6} provides spectral validation at two levels: power-iteration convergence and the eigenvalue distribution of the propagation operator: (a) the decay of the relative estimation error $|\hat{\lambda}_{\max}^{(k)} - \lambda_{\max}^{\text{true}}| / \lambda_{\max}^{\text{true}}$ with $K$; (b) the eigenvalue histogram of the propagation operator $P = I - \alpha\tilde{L}$.

The error decreases monotonically with the number of iterations: at $K = 1$ it is of order $10^{-2}$--$10^{-1}$, and at $K = 20$ it falls to about $5\times10^{-6}$, i.e., an average underestimation of $0.0005\%$ (maximum $0.003\%$), well below the $10^{-4}$ reference threshold. All 10 random aggregate distributions meet this accuracy at $K = 20$, with no tailing outliers. Fig.~\ref{fig:6}(b) reports the measured eigenvalue distribution: across the 10 aggregate distributions the eigenvalues of $P$ all lie in $[\lambda_{\min},\,\lambda_{\max}] = [1-\alpha,\,1.000] = [0.423,\,1.000]$, positive semidefinite 10/10 with a sharp peak near 1, so $\rho(P)=1\le1$; the spectral-radius constraint of Property 1 is confirmed numerically. This measurement confirms that the power iteration obtains a sufficiently accurate estimate of $\lambda_{\max}$ at $O(E)$ cost on the trimmed impact-zone graph; the training loop uses fewer iterations for efficiency, whereas the offline spectral analysis takes $K = 20$ for a stricter estimate.

\subsubsection{Total-mass conservation verification}
\label{sec:4_2_2}

The preceding analysis argued that the global total mass is approximately conserved during autoregressive rollout (within the spectral-estimation accuracy). This section gives a direct numerical verification at the operator level.

Total-mass conservation is verified at the operator level: after constructing $P = I - \alpha\tilde{L}$ on the trimmed impact-zone graph, $\|P\mathbf{1} - \mathbf{1}\|$ is directly computed; its mean over 10 random aggregate distributions is about $1.2\times10^{-8}$, i.e., the row sums of the propagation operator remain 1 to floating-point precision, and the global total mass is approximately conserved at every autoregressive step. This conservation is an algebraic property of the construction of $P$ (row-stochastic form $P\mathbf{1}=\mathbf{1}$); it does not enter the loss function and does not depend on the optimization process.

\subsection{Concrete penetration: extreme-scenario validation}
\label{sec:4_3}

The preceding two subsections validated MPNO's generality and mathematical foundations on standard PDEs and spectral analysis, respectively. This section returns to the original motivation, concrete penetration, and demonstrates MPNO's performance in an extreme engineering scenario along three dimensions: single-step accuracy, autoregressive stability, and inference efficiency. All data in this section come from the 400-sample v10 penetration dataset (\ref{sec:2_3}).

\subsubsection{Autoregressive rollout: stability comparison}
\label{sec:4_3_1}

Table~\ref{tab:4_4} reports the single-step accuracy and autoregressive rollout stability of MPNO, WNO, FNO, and MeshGraphNets on the penetration data. Among the baselines, WNO diverges to infinity on all seeds; MeshGraphNets collapses to zero prediction because of the sparse stress signal (single-step 0.9985, no learning); FNO remains stable in measurement with a single-step accuracy slightly above MPNO (0.7210 vs.\ 0.7304) but without a spectral-radius structural guarantee. MPNO achieves a single-step Rel-L2 of $0.7304\pm0.0008$ ($N=3$) with about 20K parameters. The MPNO--F13 variant additionally listed in the table recomputes the edge weights with a lightweight edge-MLP at every graph-propagation round, letting the physics coupling run through the propagation process rather than being frozen once, and improves the single-step accuracy to $0.7278\pm0.0030$ with about 37K parameters (between MPNO and FNO, still slightly below FNO) while remaining stable long-horizon. The 29-step autoregressive rollout (frames 11--39, predicted from data frame 10 onward, i.e., the entire active horizon) has bounded error on all 6 rollout test seeds across the 100/135/165 m/s cases (error about 0.98 at step 29). The rollout starts from data frame 10: in the first 10 frames the active-node fraction in the impact zone is nearly zero, providing no effective training signal; thereafter each step predicts the next frame up to frame 39, 29 steps in total. Note that single-step accuracy statistics cover all 50 test seeds of the three cases (\ref{sec:3_7_1}); the long-horizon rollout stability evaluation runs the full-horizon rollout on 6 of them (seeds 5/7/112/120/205/207, two for each of the 100/135/165 m/s evaluation velocities; 6/6 stable); the 200 m/s case serves only as training-diversity auxiliary data and is excluded from all evaluation (\ref{sec:3_7_1}); the two conventions differ and correspond respectively to single-step generalization and long-horizon stability.

This comparison agrees with the prediction of the spectral-radius dichotomy: the unconstrained baselines exhibit, across test seeds, the divergence or decay of the dichotomy (WNO's error explosion is spectral-divergence instability; MGN's prediction-energy decay is underfitting without learning, a distinct phenomenon from the spectral dichotomy); FNO remains stable in measurement, but its spectral radius is unconstrained and its stability lacks a constructive guarantee; MPNO, by contrast, constrains the spectral radius to $\rho(P) \le 1$ through power-iteration estimation and $\lambda_{\max}$ normalization, and remains stable on all test seeds.

\begin{figure}[!ht]
	\centering
	\includegraphics[width=\linewidth]{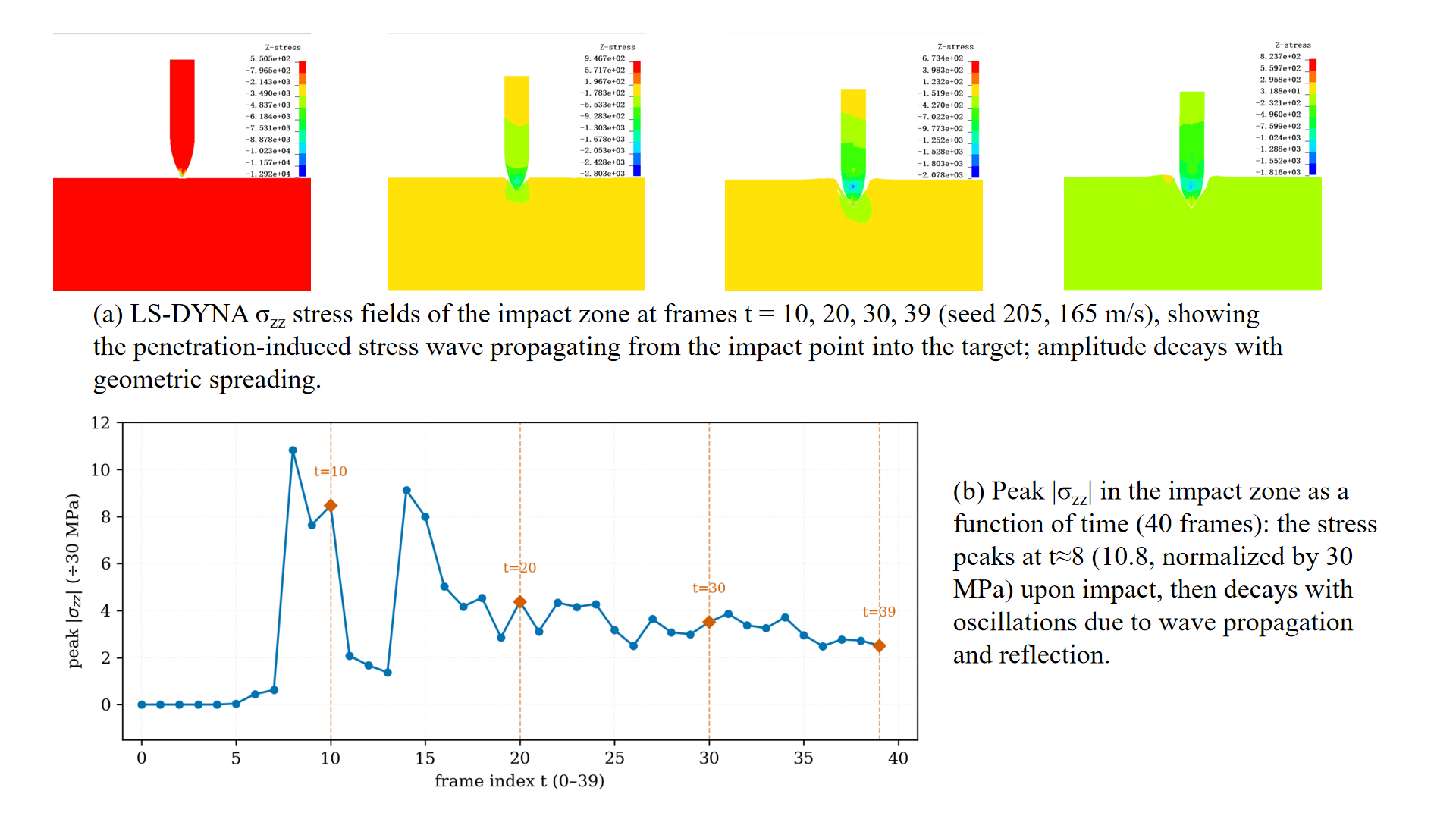}
    \caption{(a) LS-DYNA $\sigma_{zz}$ stress fields of the impact zone at frames $t=10,20,30,39$ (seed 205, 165 m/s), showing the penetration-induced stress wave propagating from the impact point; amplitude decays with geometric spreading. (b) Peak $|\sigma_{zz}|$ in the impact zone over 40 frames: the stress peaks at $t\approx8$ (10.82, normalized by 30 MPa) upon impact, then decays with oscillations due to wave propagation and reflection.}
	\label{fig:7}
\end{figure}

Fig.~\ref{fig:7} shows the spatiotemporal evolution of the impact-zone $\sigma_{zz}$ stress wave in the LS-DYNA reference solution as frames advance (test seed 205, 165 m/s, frames $t=10/20/30/39$): after the projectile enters the plate, the stress wave propagates from the impact point into the target, with the wavefront deepening per frame and the amplitude decreasing slightly with geometric spreading. It is noted that MPNO's surrogate predictions, in long-horizon autoregression, decay monotonically to a plateau in amplitude (energy-retention ratio on the order of 0.1--0.2 times the ground truth; definition in \S\ref{sec:4_1_1}). This behavior is not specific to MPNO: FNO, another single-step autoregressive surrogate, exhibits a similar dense smooth decay, whereas WNO appears as divergence in the long horizon (energy amplified severalfold, Table~\ref{tab:4_4}). The stability claim therefore refers to bounded, non-diverging error (Table~\ref{tab:4_4}, Fig.~\ref{fig:1}), not to long-horizon amplitude fidelity; single-step amplitude accuracy (Fig.~\ref{fig:7}) and long-horizon energy dissipation belong to different evaluation conventions. The distinction between benign dissipation and collapse is principled rather than a matter of endpoint amplitude: benign dissipation requires that an effective single-step map was learned, namely a single-step relative $L_2$ error well below the trivial baselines (Persistence 0.8917, Zero 1.0000, \S\ref{sec:4_3_3}), and that the long-horizon energy decays monotonically to a bounded plateau (MPNO: rollout error 0.85$\to$0.98 bounded over the 29-step active horizon, energy retention 0.1--0.2); collapse, by contrast, corresponds either to an effective map that was never learned (MGN single-step 0.9985, field collapsing toward zero amplitude) or to exponential amplification (WNO). The learnedness of the single-step map and the boundedness and monotonicity of the long-horizon decay are the operative criteria, not the endpoint amplitude alone. It is also pointed out that the absolute values of the single-step and long-horizon errors should not be compared directly with typical figures in the neural-operator literature on smooth benchmarks; the reason lies in the evaluation convention and problem characteristics, not in a method defect: the error is computed over the hard-masked active-node set (at least 5 active nodes per frame, \S\ref{sec:3_7_1}), focusing on stress-wave-dominated regions rather than a full-field average; penetration stress fields contain strong discontinuities and steep spatiotemporal gradients, the graph is small (about 200 nodes per sample), and the time series is short (40 frames), limiting the discretization resolution. The core contribution focuses on the mechanism and spectral guarantee of autoregressive stability, not on absolute accuracy.

A supplementary held-out measurement at the higher impact velocity sharpens the stability convention. On the 200 m/s test seeds of the penetration data (10 seeds, excluded from the three-case evaluation convention of \S\ref{sec:3_7_1}), the 29-step autoregressive rollout shows that FNO's apparent stability is an artifact of field collapse: its predicted field energy at the final step falls to about 0.21 of the initial value (range 0.03--0.46), whereas the reference field retains about 0.68 (range 0.10--1.06), and a relative-$L_2$ error plateauing near 1 is the signature of a prediction collapsing toward zero rather than of a bounded-error prediction. MPNO, by contrast, retains the field energy over the same rollout (1.35--2.87); it is not subject to this failure mode. A bounded relative-$L_2$ error therefore does not by itself certify stability (it cannot separate a genuinely stable rollout from a zero-field collapse), and field-energy retention is needed as a complementary criterion, the same distinction formalized above between benign dissipation and collapse. The stability contrast is also directly quantifiable: the error growth rate (the slope of the logarithm of the relative-$L_2$ error against the rollout step, fitted over the second half of the rollout) across the 6 long-horizon seeds is $+0.0005$ per step for MPNO (6/6 seeds with bounded, zero growth), $+0.0032$ per step for FNO (5/6 bounded, 1 growing), and $+0.0485$ per step for WNO (6/6 exponentially growing), so MPNO's growth rate is $6.4\times$ and $97\times$ smaller than FNO's and WNO's, respectively.

\begin{table}[!ht]
	\centering
	\caption{Single-step accuracy and autoregressive rollout comparison on the penetration data (single step with $N=3$; 29-step rollout = the entire active horizon from data frame 10)}
	\label{tab:4_4}
	\begin{tabularx}{\textwidth}{l c c X}
		\toprule
		Model & Parameters & Single-step TEST Rel-L2 & 29-step rollout \\
		\midrule
		MPNO & $\sim$20K & \textbf{$0.7304\pm0.0008$} ($N=3$) & Stable (6/6 seeds across 100/135/165 m/s; error 0.85 $\to$ 0.98 bounded) \\
		\addlinespace
		MPNO--F13 & $\sim$37K & $0.7278\pm0.0030$ ($N=3$) & Stable (6 clean seeds) \\
		\addlinespace
		WNO & $\sim$11K & $0.7422\pm0.0017$ & Divergent (to infinity on all test seeds) \\
		\addlinespace
		FNO & $\sim$79K & $0.7210\pm0.0006$ & Stable (no spectral-radius structural guarantee) \\
		\addlinespace
		MeshGraphNets & $\sim$1.0M & 0.9985 (no learning) & Prediction energy decays with rollout \\
		\addlinespace
		\multicolumn{4}{@{}p{\linewidth}@{}}{\small Note: MPNO--F13 is an improved variant of MPNO that recomputes the edge weights with an edge-MLP at every graph-propagation round (\S\ref{sec:4_3_1}), letting the physics coupling run through the propagation process rather than being frozen once; with about 37K parameters (about half of FNO) it raises the single-step accuracy to 0.7278, narrowing the gap to FNO to $\pm 0.007$, while remaining stable long-horizon.} \\
		\bottomrule
	\end{tabularx}
\end{table}

MGN \cite{ref9} shares the graph topology with MPNO but learns the message function with an edge-wise MLP, without a global spectral constraint; its single-step 0.9985 fails to learn an effective map and its prediction energy decays with rollout, consistent with the expectation that ``edge-wise message passing without a spectral constraint is hard to train on extremely sparse data.''

\subsubsection{Physics-coupling effect: diversity of edge-weight information}
\label{sec:4_3_2}

The measured statistics of the edge weights reveal a phenomenon counter to intuition: before training (pure physics skeleton $T\cdot|\mathbf{t}|$, MLP weights zeroed) the edge weights are large in scale (standard deviation 0.3254, maximum 6.858, 100\% nonzero); after training (MLP active) the weights are compressed into a sparse heavy-tailed distribution (standard deviation $5.89\times10^{-4}$, maximum about 0.025, median near 0), with the maximum compressed about 275$\times$ and the standard deviation about 553$\times$, and the coefficient of variation (CV) rising from 4.6 to 8.5. That is, the physics skeleton provides large initial weights based on the traction magnitude $|\mathbf{t}|$, and the MLP correction learned during training compresses them; the coupling acts mainly as suppression/selection rather than amplification: the vast majority of edges are suppressed to near zero (blocking noise propagation), while a small number of key force-transfer paths are retained.

\subsubsection{Trivial-baseline comparison: has MPNO learned effective temporal evolution}
\label{sec:4_3_3}

Since the single-step errors of all neural operators cluster near 0.72--0.74, it is necessary to confirm that MPNO's predictions are not the trivial solution of ``approximately copying the previous frame.'' To this end, two trivial baselines are introduced (same test set, same split and mask): Persistence (predicting $\hat{\boldsymbol{\sigma}}_{t+1}=\boldsymbol{\sigma}_t$, i.e., ``same as the previous frame'') and Zero (predicting all zeros). The evaluation convention matches the main task: the mask $\mathcal{A} = \{i : \|\boldsymbol{\sigma}_i^{\text{GT}}\|_2 > 0.01\}$ ($\tau=0.01$, normalized stress) defines the active nodes, and a frame enters the statistics only if it has at least 5 active nodes (\S\ref{sec:3_7_1}); the candidate-frame statistics also appear in \S\ref{sec:3_7_1}. The results are: MPNO single-step Rel-L2 $0.7304\pm0.0008$, Persistence 0.8917, Zero 1.0000 (three-case convention, 50 test seeds); MPNO improves over Persistence by 18.1\% and over Zero by 27.0\%, showing that the model has learned the temporal evolution of the stress field rather than degenerating into an identity map. It is worth noting that Persistence degrades markedly under the three-case convention excluding 200 m/s (0.8917): the 200 m/s case evolves slowly between frames and is most easily approximated by the previous frame, whereas under the three impact conditions of the evaluation, this trivial baseline is stronger and MPNO's advantage over it is correspondingly larger. This comparison also explains why the single-step accuracy of all neural operators clusters at 0.72--0.74: the stress field evolves slowly frame by frame, Persistence is itself a strong baseline, and single-step accuracy is largely determined by fitting the slowly varying field.

\subsection{Spectral-stability mechanism and component ablations}
\label{sec:4_4}

This section examines MPNO's stability guarantee at the mechanism level: \S\ref{sec:4_4_1} measures the spectral properties of the propagation operator and the full model on the trimmed impact-zone graph; \S\ref{sec:4_4_2} removes components one by one and quantifies each component's independent contribution; \S\ref{sec:4_4_3} reports inference efficiency.

\subsubsection{Spectral-radius measurements and the mechanism of action}
\label{sec:4_4_1}

\begin{figure}[!ht]
	\centering
	\includegraphics[width=\linewidth]{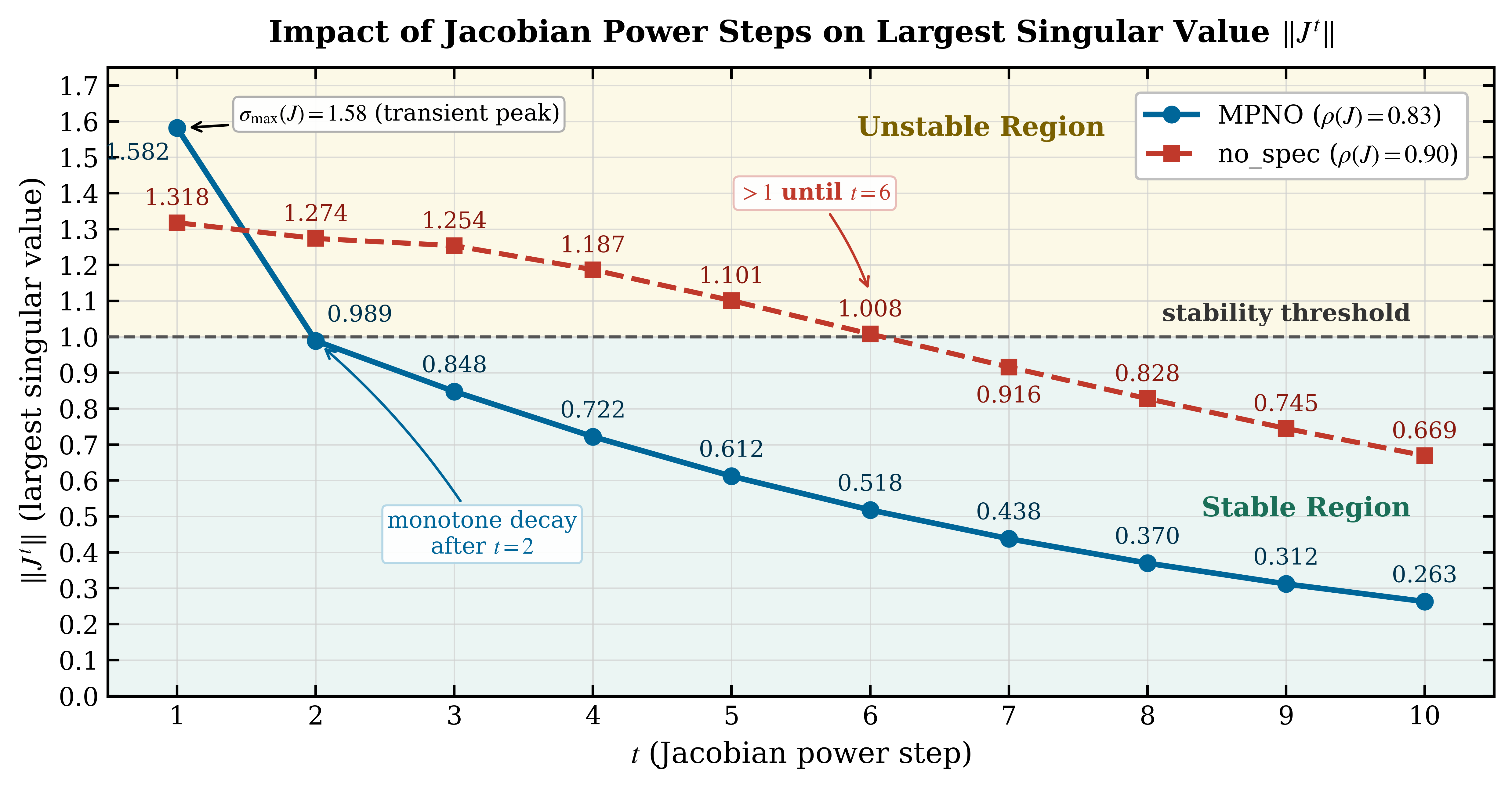}
	\caption{Pseudospectral transient growth of the full-model Jacobian (seed 55, frame 10, $N=224$). $\|J^t\|$ (largest singular value of the $t$-step Jacobian power) is shown for MPNO (solid blue) and no\_spec (dashed red); the horizontal dashed line marks the stability threshold $\|J^t\|=1$. MPNO's transient peaks at $t=1$ (1.58, equal to $\sigma_{\max}(J)=1.58$) then collapses below 1 at $t=2$ and decays monotonically to 0.263, reconciling $\sigma_{\max}(J)>1$ with asymptotic contraction at this operating point ($\rho(J)=0.83\le1$). no\_spec retains growth above 1 through $t=6$ ($\rho(J)=0.90$). Model spectral radii $\rho(J)$ are positioned on the $\rho$-axis taxonomy in Fig.~\ref{fig:4}.}
	\label{fig:8}
\end{figure}

\begin{figure}[!ht]
	\centering
	\includegraphics[width=\linewidth]{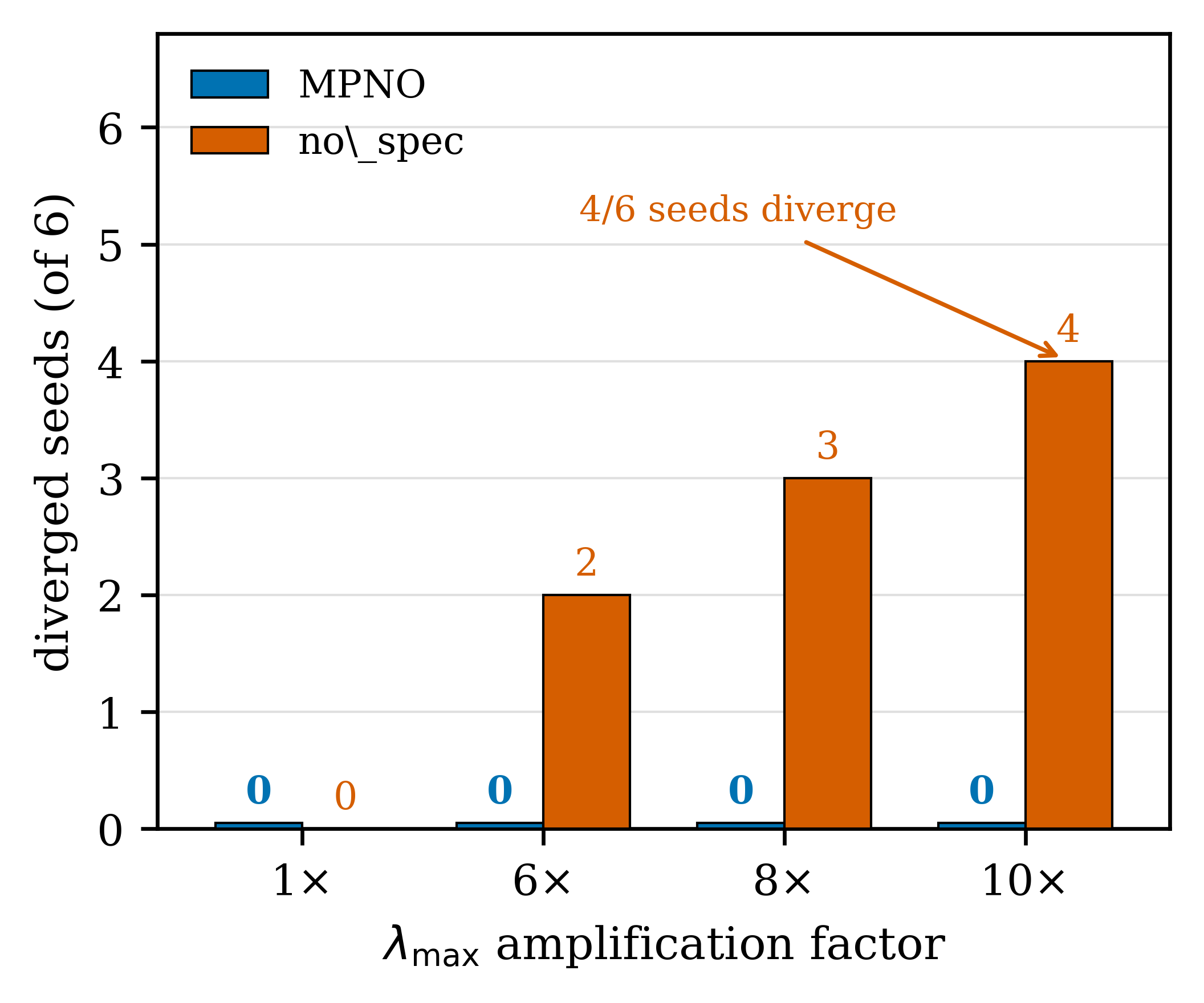}
	\caption{Controlled counterfactual: inflating the learned $\lambda_{\max}$. The estimated $\lambda_{\max}$ is multiplied by $1\times/6\times/8\times/10\times$ before re-normalizing the propagation operator (all else unchanged). MPNO (blue) preserves its stability plateau at every amplification factor (0/6 seeds diverge); no\_spec (red) loses emergent stability monotonically, diverging in 2/6, 3/6, and 4/6 seeds at $6\times$, $8\times$, and $10\times$. The divergence of no\_spec under spectrum inflation, despite identical stability on nominal data, shows that spectral normalization, not incidental training, underlies MPNO's stability.}
	\label{fig:9}
\end{figure}

To clarify the mechanism of the spectral constraint, the spectral properties of the propagation operator are measured on the trimmed impact-zone graph (seed 55, frame 10, reproducible). At this representative operating point, MPNO's propagation rate is $\alpha=0.577$ with $\lambda_{\max}(L)=0.025$ under the MPNO edge-weight convention, giving $\rho(P)=1.000$; across the 6 clean seeds (frame-10 operating points), $\alpha\lambda_{\max}=0.002$--$0.076$ and $\rho(P)=1.000$: the spectral radius is decoupled from the scale of $\lambda_{\max}$ and from the optimization trajectory. In contrast, no\_spec (normalization removed) learns edge-weight magnitudes about 90 times those of MPNO, and its linear operator on the training data has $\alpha\lambda_{\max}=4.3$--$7.0$, already exceeding the linear stability bound of 2; the bounded output that no\_spec exhibits in routine rollout is not a structural guarantee of the linear propagation operator but the result of emergent clamping by the LayerNorm nonlinearity. The $\lambda_{\max}$ extrapolation experiment (inflating $\lambda_{\max}$ at inference by scaling all edge weights, without changing the input features) provides the key contrast between the two mechanisms: because MPNO recomputes $\lambda_{\max}$ per graph through spectral normalization, its output values on the 6 test seeds are unchanged under a 10$\times$ inflation of $\lambda_{\max}$ (maximum deviation about $\sim10^{-6}$); under a 6--10$\times$ inflation, no\_spec's diverging seeds increase monotonically with the amplification factor (0/6 $\to$ 2/6 $\to$ 3/6 $\to$ 4/6), and the non-diverging seeds also show chaotic oscillation. This contrast shows that the value of the spectral constraint is not an empirical necessity at the current data scale, but a lift of stability from the chance of the data/training trajectory to a designable, dependable architectural property. In addition, data-driven emergent contraction is not unique to MPNO: the POD-DMD linear reduced-basis model (linear ROM), without any spectral constraint, emerges $\rho<1$ contraction merely from linear fitting of the data (measured about 0.88--0.97 per velocity case) and remains rollout-stable on the same set of clean test seeds, but its single-step accuracy (about 0.78) is below MPNO's (0.7304) and it is limited by the expressivity of a linear subspace, whereas MPNO, with nonlinear representation, approximates nonlinear wave dynamics more closely under the same stability guarantee. The full-model Jacobian spectral radius is obtained by explicitly constructing $J$ (autograd) and taking the largest-modulus eigenvalue (seed 55, frame 10, a representative operating point): MPNO $\rho(J)=0.83$, no\_spec $\rho(J)=0.90$; under the same convention FNO $\rho(J)=0.92$ (right at the critical line; its stability is emergent from training, without a constructive guarantee) and WNO $\rho(J)=1.64>1$ (spectral radius unconstrained, consistent with the dichotomy prediction and directly corresponding to its autoregressive divergence). Across repeated measurements on the 6 clean rollout seeds, the empirical full-model spectral radius stays in a narrow band above the critical boundary, with MPNO ($\approx1.05$) closer to $\rho=1$ than no\_spec ($\approx1.13$), the ordering (MPNO $<$ no\_spec) holding stably across seeds; the mechanism systematically pulls the empirical spectral radius toward the theoretical critical boundary, the residual excess reflecting nonlinear corrections to the linearized operator (limitation (i), \S\ref{sec:4_5_2}). The local linearized spectral radius of the single-step operator is not directly the asymptotic behavior of autoregression; because the full Jacobian is nonnormal, its spectral norm $\sigma_{\max}(J)$ is measured above 1 (MPNO $=1.58$, no\_spec $=1.32$), corresponding to pseudospectral transient growth over finite steps (MPNO's $\|J^t\|$ rises to 1.58 at the first step and then falls, below 1 from $t\ge2$ onward and monotonically decreasing \cite{ref47}), which does not contradict asymptotic non-divergence. The operator-level spectral guarantee ($\rho(P)\le1$) and the full-model rollout measurement together form two layers of boundary: the former is a constructive guarantee independent of the training trajectory, and the latter is confirmed by the autoregressive rollout (\S\ref{sec:4_3}). Fig.~\ref{fig:8} plots the pseudospectral transient growth of the full-model Jacobian $\|J^t\|$ versus rollout step, and Fig.~\ref{fig:9} contrasts the diverging-seed counts of the $\lambda_{\max}$ extrapolation stress test.

\subsubsection{Component ablations}
\label{sec:4_4_2}

To isolate each component's independent contribution, components are removed one by one from the full model, and each configuration is trained and evaluated independently under the same setting (SEED=42, single-step TEST Rel-L2, masking convention identical to the comparison table, i.e., $\tau=0.01$ and $\ge$5 active nodes per frame). Table~\ref{tab:4_7} reports the ablation results.

\begin{table}[!ht]
	\centering
	\footnotesize
	\setlength{\tabcolsep}{4pt}
	\caption{Ablation study (single-step TEST Rel-L2; main baseline is the $N=3$ mean, ablations are single runs with SEED=42)}
	\label{tab:4_7}
	\begin{tabularx}{\textwidth}{l p{2.6cm} c c >{\raggedright\arraybackslash}X}
		\toprule
		Configuration & \makecell[l]{Removed\\component} & \makecell[c]{Three-case\\single-step\\Rel-L2} & \makecell[c]{All-case\\single-step\\Rel-L2} & Conclusion \\
		\midrule
		baseline       & (full model)             & $0.7304\pm0.0008$ & 0.7108 & Baseline ($N=3$) \\
		no\_spec       & Spectral normalization   & 0.7366            & 0.7135 & Consistent degradation across conventions \\
		no\_phys       & Physics-coupled edge weight & 0.7291        & 0.7256 & Three-case neutral; benefits high-velocity cases \\
		no\_round\_mlp & round\_mlps cyclic post-processing & 0.7314  & 0.7201 & Three-case neutral; useful at high velocity \\
		no\_speed      & Velocity coupling        & --                & 0.7175 & Useful \\
		no\_appnp      & APPNP residual connection & --               & 0.7191 & Useful \\
		no\_force\_gate& Traction-magnitude gate  & --                & 0.7135 & Minor effect \\
		no\_ln         & LayerNorm                & --                & 0.7115 & Nearly no effect \\
		no\_hard\_mask & Hard-masked MSE          & --                & 0.7114 & Nearly no effect \\
		\bottomrule
	\end{tabularx}
	\par\vspace{0.3em}
	\noindent\scriptsize\textit{Note:} The three-case column is the primary evaluation convention: baseline is the $N=3$ mean ($0.7304\pm0.0008$), ablations are single runs with SEED=42; the all-case column (incl.\ 200 m/s) is a single-run SEED=42 development-phase evaluation, and -- indicates that the configuration was not separately measured under that convention. All-case effect sizes are reported rounded to three decimals; the all-case comparison column shows the component benefit under high-velocity impact; ablation differences of $\pm0.002$ lie within seed-to-seed variance.
\end{table}

Table~\ref{tab:4_7} shows that spectral normalization (no\_spec) is the only component that degrades consistently across conventions: removing it increases the single-step error by +0.006 under the three-case convention, in the same direction as the all-case comparison including 200 m/s (+0.003), indicating that the contribution of the spectral-stability mechanism is an empirical result independent of the convention. The physics coupling (no\_phys) and round\_mlps are accuracy-neutral under the three-case convention ($-$0.001 / +0.001, within the seed-to-seed noise band), with their accuracy benefit concentrated in high-velocity impact: in the all-case comparison, removing no\_phys increases the error by +0.015 and no\_round\_mlp by +0.009. The removal effects of the remaining components (APPNP residual, velocity coupling, traction gate, LayerNorm, hard-masked MSE) all stay within $+0.008$ in the all-case development-phase evaluation, a minor contribution.

A 20-step rollout across all ablation configurations shows that the model remains stable (bounded error, no collapse) even with every component removed: the architecture is robust to single-component removal (single-step degradation all $\le$0.015), and stability is a holistic architectural property rather than the irreplaceability of any single component. The accuracy contribution of the physics coupling and round\_mlps concentrates in high-velocity impact (all-case $+0.015$/$+0.009$), while the spectral mechanism is the only component whose removal degrades every evaluation convention (no\_spec $+0.006$/$+0.003$). Consistent with this robustness, single-step accuracy is largely carried by the encoder--decoder fitting of the slowly varying field, and the propagation components deliver the stability guarantee rather than incremental single-step accuracy. no\_spec's bounded output is emergent nonlinearity rather than a structural guarantee of the linear propagation operator; its measured spectral behavior and the breakdown of emergent stability under $\lambda_{\max}$ extrapolation are detailed in \S\ref{sec:4_4_1}; the value of spectral normalization is thereby delimited as: decoupling $\alpha$ from $\lambda_{\max}$ and making the constructive $\rho(P)\le1$ independent of the training trajectory.

\textbf{Honest account of the statistical convention.} Note that the ablation results of Table~\ref{tab:4_7} are single runs with fixed SEED=42, whereas the main-task baseline is the mean $\pm$ standard deviation of $N=3$ independent initializations ($0.7304\pm0.0008$). Among the ablation effect sizes ($-$0.001 to +0.006), the $+0.001\sim+0.006$ effects are of the same magnitude as the seed-to-seed fluctuation of a single run and should not be interpreted as definite causal differences; no\_spec provides evidence of degradation consistent across conventions (three-case +0.006 / all-case +0.003, same direction), and the accuracy effects of the physics coupling and round\_mlps fall within the noise band under the three-case convention, with their benefit appearing in the all-case comparison (high-velocity cases). The ablation conclusions should therefore be read as ``a relative importance ranking of the components'' and ``cross-convention robustness of the spectral-normalization contribution,'' rather than ``precise effect sizes of the components.''

\textbf{Role of the physics coupling.} The accuracy-neutrality of the physics coupling under the three-case convention does not contradict its claimed contribution: single-step accuracy on this slowly varying field is largely carried by the encoder--decoder, whereas the physics coupling is retained for its cross-scenario transferability of the edge-weight formula (\S\ref{sec:4_1}), its measurable benefit at high impact velocity (all-case $+0.015$), and its zero-parameter contribution to parameter efficiency; no generic three-case accuracy gain is claimed. The conclusion should likewise not be read as the physics coupling carrying the stability mechanism; that is the role of spectral normalization, the only component whose removal degrades every evaluation convention (no\_spec $+0.006$/$+0.003$); the coupling contributes to the edge-weight construction and its transfer, not to the spectral guarantee itself.

\subsubsection{Inference efficiency}
\label{sec:4_4_3}

\begin{figure}[!ht]
	\centering
	\includegraphics[width=\linewidth]{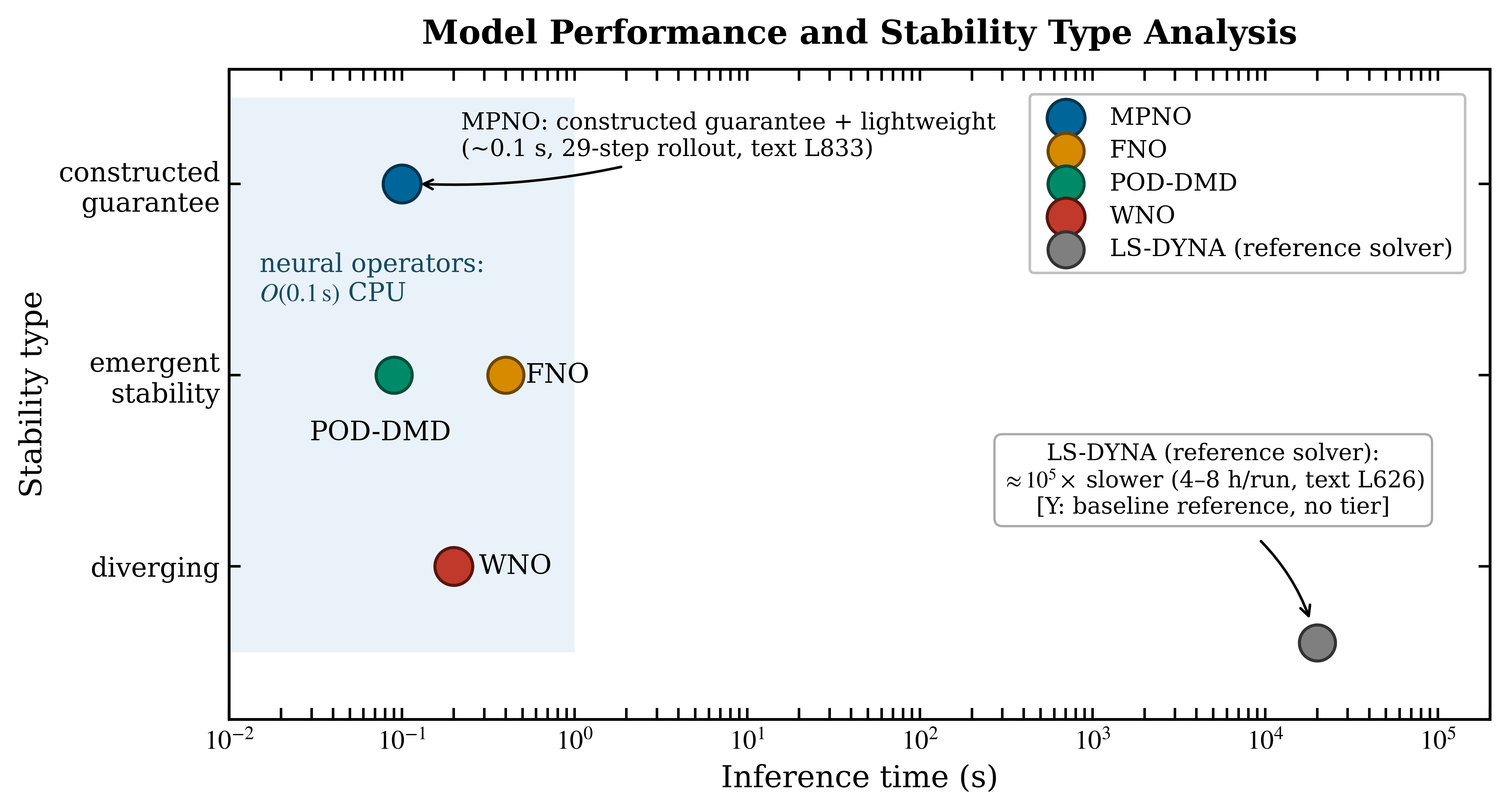}
	\caption{Efficiency and stability relative to baselines. Horizontal axis: inference time per 29-step rollout (log scale); vertical axis: stability type (constructed / emergent / diverging). MPNO is the only model combining a constructed stability guarantee with lightweight inference ($\sim$0.1~s on CPU vs.\ 4--8~h for the LS-DYNA reference, a $\approx$10$^5\times$ speed-up); FNO is stable but without a guarantee; WNO diverges; POD-DMD is emergent. Parameter counts are listed in Table \ref{tab:params}.}
	\label{fig:10}
\end{figure}

MPNO's single-step inference takes about 2.4 ms on CPU and about 3.7 ms on GPU (with such a small model, GPU kernel-launch overhead dominates and CPU is actually faster); the autoregressive rollout covering the entire data window (29 steps) takes about 100 ms, a speed-up of roughly $10^5$ over LS-DYNA. Fig.~\ref{fig:10} locates MPNO's uniqueness in the ``constructive guarantee + lightweight inference'' combination by comparing inference time and stability type. Since the single-step errors are similar across the three neural operators (Table~\ref{tab:4_4}), the combination of efficiency difference and stability-guarantee type is the main dimension distinguishing the methods.

\subsection{Cross-domain transient generalization: compressible Euler}
\label{sec:repro_euler}

The stability advantage established on the penetration dataset is not restricted to a single problem. To verify that the spectral guarantee transfers to genuinely transient dynamics outside the training domain, the operator-learning setting is reproduced on the two-dimensional compressible Euler equations. Training data are generated locally with the official PDEBench~\cite{ref30} generation pipeline (a JAX HLLC finite-volume solver) under the compressible-Euler random-initialization configuration ($M_0 = 0.1$, $\eta = \zeta = 0.01$, periodic boundaries, $128\times128$ grid), yielding 112 trajectories of 21 frames ($\rho,u,v,p$); the out-of-distribution test set is the official PDEBench Sod shock-tube data (a single trajectory of 101 frames at $1024^2$ resolution, downsampled to $64^2$). The lightweight MPNO used here matches the uniform-weight variant of the Burgers/Darcy benchmarks: uniform edge weights, the Markov propagation kernel $P = I - \alpha\tilde{L}$ with the spectral-radius constraint of Property~1, and a four-channel decoder for $(\rho,u,v,p)$, since the physics-coupled edge weights of Section~\ref{sec:3_4} do not apply to a gas. All four models share the identical protocol (resolution 64, initial step 10, 200 epochs, batch size 16, AdamW with cosine annealing), and evaluation is a global relative-$L_2$ error over the full domain without an active mask.

Table~\ref{tab:4_9} reports single-step accuracy and long-horizon autoregressive stability on the cross-domain shock-tube rollout. MPNO's single-step error ($8.82 \pm 3.05$) trails the two spectral baselines ($60\times$ and $4\times$ larger in parameters) and improves on MeshGraphNet. The single-step ranking is secondary here: all four errors are inflated by the cross-domain scale shift documented below, so the internally valid comparison is long-horizon stability, where the contrast is stark: MPNO and MeshGraphNet complete all 91 autoregressive steps without divergence, whereas FNO diverges at step 57 and WNO at step 4; MPNO additionally attains the lowest final-step error (17.96 vs.\ 22.74 for MeshGraphNet). The spectral guarantee thus transfers: on a genuinely transient shock problem outside the training distribution, the architecture whose stability is a construction, rather than emergent or absent, retains bounded-error autoregressive rollout. Quantifying the stability contrast, the error growth rate (the slope of the logarithm of the relative-$L_2$ error against the rollout step, fitted over the second half of the rollout) is $-0.0014$ per step for MPNO over the full 91-step horizon (the only negative growth rate in the comparison, i.e., the only trajectory whose error contracts with step), whereas FNO exhibits $+0.049$ per step before diverging at step 57 and WNO diverges at step 4. Two caveats apply. The OOD test is a single shock-tube trajectory rather than a multi-sample statistic. And the single-step errors are elevated for all four models: the shock-tube density range and field magnitudes lie outside the training distribution, and the dimensionless relative-$L_2$ ratio is computed over near-vacuum regions with small ground-truth norms. The shift is identical across models, so the comparison remains internally valid. The transfer established here is one of stability, not accuracy.

\begin{table}[!ht]
	\centering
	\caption{Cross-domain out-of-distribution generalization on the PDEBench 2D compressible-Euler Sod shock-tube (single trajectory, 101 frames; all models share one protocol). rel.-L2 = dimensionless relative $L_2$ ratio $\|\hat{\mathbf{u}}-\mathbf{u}\|_F/\|\mathbf{u}\|_F$; ``diverged'' = the error exceeds 50 or NaN before the rollout ends.}
	\label{tab:4_9}
	\begin{tabularx}{\textwidth}{l c c c c X}
		\toprule
		Model & \#Params & \makecell[c]{Single-step\\rel.-L2 ($\times1$)} & \makecell[c]{Long-horizon\\completed} & Diverged & Final-step error \\
		\midrule
		MPNO (lightweight) & 19{,}973 & $8.82\pm3.05$ & 91/91 & No & 17.96 \\
		\addlinespace
		FNO & 1{,}187{,}556 & $6.67\pm2.73$ & 57/91 & Yes & $>50$ \\
		\addlinespace
		WNO & 73{,}956 & $6.68\pm4.29$ & 4/91 & Yes & $>50$ \\
		\addlinespace
		MeshGraphNet & 63{,}780 & $11.79\pm3.49$ & 91/91 & No & 22.74 \\
		\bottomrule
	\end{tabularx}
\end{table}

\section{Discussion and limitations}
\label{sec:4_5}

The experiments in \ref{sec:4_1} through \ref{sec:4_3} validate MPNO's core claims at three levels (standard PDEs, spectral analysis, and extreme engineering scenarios): autoregressive stability is constructively guaranteed by the spectral constraint of the propagation matrix (power-iteration spectral estimation + $\lambda_{\max}$ normalization), with generality independent of the PDE type. This section delimits the applicability boundary of the methodology on this basis and candidly states its limitations, giving the reader a complete picture of ``when it applies and when it does not.''

\subsection{Applicability conditions}
\label{sec:4_5_1}

MPNO's strategy embeds the physics prior (constitutive impedance) into the propagation structure through the edge-weight construction and constructively guarantees stability through the algebraic spectral constraint. This strategy takes effect under three mutually supporting conditions, corresponding respectively to the three design choices. First, the system has a natural graph topology. Any system whose solution domain is discretized by a mesh provides cell--contact-face adjacency; the six-face adjacency of hexahedral meshes yields a regular graph structure (\ref{sec:3_1}). For unstructured meshes or particle systems (SPH, DEM), adjacency is rebuilt from spatial proximity or contact, and the spectral conclusions (Property 1 does not depend on the edge-set definition) still hold. Second, the evolution is governed by local conservation laws. The Laplacian filter $P = I - \alpha\tilde{L}$ presumes that field variables evolve smoothly in space and conserve total mass, which covers the vast majority of transient-dynamics problems governed by mass, momentum, and energy conservation; for diffusion--reaction systems with sources/sinks, a source term must be appended, in which case the total-mass conservation guarantee fails but the spectral-radius constraint of Property 1 still holds. Third, the evolution can be approximated as a Markov process. The state update depends only on the current state, which holds naturally for short-range-correlated systems; for rate-dependent constitutive models with strong historical memory, the Markov approximation is compensated under high-strain-rate penetration ($10^3$--$10^5$ s$^{-1}$) by the strain-rate and acceleration information implicit in the state update (\ref{sec:3_1_2}).

The replaceability of the physics-coupling formula constitutes a fourth piece of evidence for applicability. The acoustic-impedance harmonic mean $T_{ij}=2Z_iZ_j/(Z_i+Z_j)$ (penetration), the harmonic-mean permeability $w_{ij}=2K_iK_j/(K_i+K_j)$ (Darcy), and the uniform weight (Burgers) share the same architecture, differing only in the material-property variable. The framework is not bound to any specific physical formula but to a broader condition: the physics coupling can be expressed by node-local variables. This lets MPNO transfer to new scenarios by replacing the local material-property variable.

\subsection{Limitations}
\label{sec:4_5_2}

(i) The spectral guarantee is limited to the linear propagation operator. The spectral-radius constraint $\rho(P)\le1$ of Property 1 holds for the linear propagation operator $P$, bounding the autoregressive error at the operator level; however, the full model is a nonlinear composition of the encoder--decoder MLPs with $P$, and its long-horizon stability is not directly guaranteed by the spectral constraint but confirmed empirically by the rollout curve (Fig.~\ref{fig:1}). The spectral constraint also does not constitute an accuracy guarantee: it bounds the error but does not constrain its size, which is decided by the encoder--decoder expressivity and training quality (the accuracy in Table~\ref{tab:4_4} is a measured post-convergence value, not a direct corollary of the spectral constraint); equating ``spectral stability'' with ``arbitrary accuracy,'' or treating the stability advantage as an exemption from the accuracy bottleneck, would both be misreadings. Extending the spectral constraint to the nonlinear composition (e.g., applying spectral normalization to the encoder/decoder to tighten the Lipschitz constant) is left for future work.

(ii) Data and resolution boundaries. The current penetration experiments are based on 400 samples, four initial velocities (100/135/165/200 m/s), and a 10mm mesh, which can capture the macroscopic low-frequency response of impact; high-frequency oscillations at higher resolution (material fracture, local damage patterns) lie outside the data range considered. Two further technical conditions hold within this scope: the physics skeleton is assumed to be approximately consistent with the material behavior, since the learned correction is local to the skeleton; and the power-iteration spectral estimate assumes a non-degenerate spectral gap, which is satisfied at the 10mm resolution.

The above limitations are an honest delimitation of the applicability domain, not a rejection of the methodology. MPNO's contribution is not to replace all neural-operator paradigms, but to point out a previously overlooked design path: when stability must be an architectural property rather than a loss-function objective, the spectral structure of the graph propagation matrix provides a ready-made algebraic tool.

\section{Conclusions}
\label{sec:5}

The Markov physics-informed neural operator (MPNO) is proposed as a graph-structured neural-operator architecture that reformulates transient-dynamics PDEs as a Markov propagation process on a spatially discretized graph. Its core methodology is defined by a chain of spectral constraints: the physical construction of the adjacency matrix $W$ $\to$ the spectral properties of the graph Laplacian $L = D - W$ $\to$ the Rayleigh-quotient power iteration providing an online estimate of $\lambda_{\max}$ $\to$ the spectral radius of the normalized propagation operator $P = I - \alpha\tilde{L}$ constrained to $\rho(P) \le 1$. These spectral properties (embedded in the architecture), together with numerical experiments on three PDE scenarios (Burgers shocks, Darcy heterogeneous flow, and high-velocity concrete penetration), support the core claim: \textbf{autoregressive stability cannot be reliably optimized as a target of the loss function; it should be explicitly constructed as a spectral property of the architecture.}

The experimental results show that: MPNO achieves a single-step Rel-L2 of $0.7304\pm0.0008$ on the concrete-penetration scenario with about 20K parameters (better than WNO, comparable to FNO's single-step error, at about one quarter of FNO's parameters), and its 29-step autoregressive rollout (the entire active horizon) is bounded-error and stable across the 100/135/165 m/s cases (\S\ref{sec:4_3}); on Burgers, 30 steps of autoregression do not collapse, and on Darcy, replacing the physics-coupling variable verifies the cross-scenario transferability of the edge-weight formula (\S\ref{sec:4_1}). Among the baselines' instabilities on the test seeds, WNO's divergence falls within the dissipation/divergence dichotomy characterized by the spectral-radius dichotomy, MGN's zero prediction is underfitting without learning (a distinct phenomenon from the spectral dichotomy), and FNO, though measured stable, lacks a constructive guarantee. The causal role of spectral normalization is isolated by the counterfactual experiment of Fig.~\ref{fig:9}: inflating $\lambda_{\max}$ by 6/8/10$\times$ at inference monotonically breaks no\_spec's emergent stability (diverging seeds 2/6$\to$3/6$\to$4/6), whereas MPNO's output values are unchanged because it recomputes $\lambda_{\max}$ per graph (to $\sim10^{-6}$), showing that stability is guaranteed by construction rather than by the chance of the training trajectory. The proposed method applies to graph-topology systems governed by local conservation laws whose evolution can be approximated as a Markov process (\S\ref{sec:4_5_1} delimits the applicability conditions). MPNO points to a design principle: when physics priors are embedded in the architecture in a constructive form, the data-driven components can remain lightweight.

\section*{Conflict of interest statement}
The authors declare that they have no known competing financial interests or personal relationships that could have appeared to influence the work reported.

\section*{Data availability statement}
All data are generated by the procedures described in Section~\ref{sec:2} and are independently reproducible: Burgers from the Cole--Hopf analytic solution (\S\ref{sec:2_1}), Darcy from an FFT Gaussian random field and the five-point finite-difference method (\S\ref{sec:2_2}), and the concrete-penetration data by LS-DYNA/Explicit simulations (RSA aggregate packing, JH-2/CSCM constitutive models; \S\ref{sec:2_3} and Table~\ref{tab:1}); the generation scripts accompany the code. The penetration dataset (stress time series and aggregate volume-fraction fields of 400 cases) will be made available by the authors upon reasonable request.

\section*{Code availability and reproducibility}
MPNO is implemented in PyTorch. All key experimental settings for reproducibility are provided: the data split (70:15:15, SEED=42), the seed strategy ($N=3$ baseline initializations, single-seed ablations), the evaluation convention (single-step Rel-L2, $\tau=0.01$, $\ge5$ active nodes per frame), per-scenario training hyperparameters, and the deterministic long-horizon seed rule (\S\ref{sec:3_7_1}, \S\ref{sec:4_1}--\S\ref{sec:4_4}). The codebase (data-generation and resampling scripts, training/evaluation entry points, the \texttt{torch.no\_grad()} power-iteration spectral normalization, and ablation switches) will be made available by the authors upon reasonable request.

\end{document}